\pdfoutput=1

\documentclass[11pt]{article}

\usepackage{EMNLP2023}

\usepackage{lipsum} 

\usepackage{times}
\usepackage{latexsym}
\usepackage{makecell}
\usepackage{hyperref}
\usepackage{tablefootnote}
\usepackage{threeparttable}
\usepackage{makecell}

\usepackage{tabularx}
\usepackage{mathtools, amssymb}
\usepackage{xstring}
\usepackage{float,subcaption,stfloats}
\usepackage{booktabs,multirow}
\usepackage{graphicx,tikz}
\usepackage{enumitem,cleveref}
\usepackage[noend]{algpseudocode}
\usepackage{algorithm, algorithmicx}

\usepackage{amsfonts,amssymb,amsmath,amsthm,enumitem}
\usepackage{fontawesome}
\usepackage[most]{tcolorbox}

\usepackage{xcolor,soul}
\usepackage{colortbl}

\definecolor{Maroon}{cmyk}{0, 0.87, 0.68, 0.32}
\definecolor{BrickRed}{cmyk}{0.00,0.74,0.87,0.23}
\definecolor{RawSienna}{cmyk}{0.00,0.62,0.91,0.38}
\definecolor{Mahogany}{cmyk}{0.00,0.72,0.83,0.29}
\definecolor{Sepia}{cmyk}{0.00,0.76,0.99,0.64}
\definecolor{lightgrey}{rgb}{0.875, 0.875, 0.875}
\definecolor{darkgreen}{rgb}{0.01, 0.75, 0.24}
\definecolor{lightgreen}{rgb}{0.63, 0.85, 0.61}
\definecolor{lightred}{rgb}{0.95, 0.59, 0.61}
\definecolor{csplit}{rgb}{0.45, 0.45, 0.45}

\definecolor{clabelbox}{rgb}{0.25, 0.25, 0.25}

\usepackage{mathabx}

\definecolor{beautifulblue}{RGB}{0,76,153}
\definecolor{myblue}{RGB}{215,238,247}
\definecolor{caddback}{rgb}{0.90, 0.98, 0.96}
\definecolor{cadd}{rgb}{0, 0.47, 0.34}
\definecolor{cdelback}{rgb}{1, 0.94, 0.92}
\definecolor{cdel}{rgb}{0.83, 0.32, 0.16}

\expandafter\def\expandafter\normalsize\expandafter{%
    \normalsize%
    \setlength\abovedisplayskip{3pt}%
    \setlength\belowdisplayskip{4pt}%
    \setlength\abovedisplayshortskip{-8pt}%
    \setlength\belowdisplayshortskip{2pt}%
}

\newcommand{\dashedline}{
  \begin{center}
    \tikz \draw [dashed] (0,0) -- (\linewidth,0);
  \end{center}
}

\newcommand{\RomanNumeralCaps}[1]{\MakeUppercase{\romannumeral #1}}

\usepackage{xifthen}
\usetikzlibrary{positioning}
\usepackage{soul}

\definecolor{modelchat}{rgb}{0.91,0.91,0.91}
\definecolor{userchat}{rgb}{0.55,0.73,0.96}
\definecolor{sessionchat}{rgb}{0.2,0.2,0.2}

\NewDocumentEnvironment{chat}{ o }{%
   \IfNoValueTF{#1} { \begin{figure} }{ \begin{figure}[#1] }
}{%
   \end{figure}
}

\newcommand{\session}[1]{%
       \tikz{
       \draw[anchor=north west,draw=sessionchat, text=sessionchat] (0,0) -- (\linewidth,0) node [midway, above] {\small{#1}};}%
       \vspace{2pt}%
}
\newcommand{\model}[2][]{%
       \tikz{%
        \node[anchor=north west,fill=userchat,text width=0.92\linewidth,rounded corners=2pt,align=left] (content) {\strut#2\strut};%
        \ifthenelse{\isempty{#1}}{}{\node[below=-2pt of content,align=left,text width=0.92\linewidth]{\footnotesize{#1}};}%
       }%
       \ifthenelse{\isempty{#1}}{\vspace{2pt}}{\vspace{-2pt}}%
}
\newcommand{\user}[2][]{%
       \tikz{%
       \node[anchor=north west] {};
       \node[anchor=north west,fill=modelchat,right=0.05\linewidth,text width=0.92\linewidth,rounded corners=2pt] (content) {\strut#2\strut};%
       \ifthenelse{\isempty{#1}}{}{\node[below=-2pt of content,align=right,text width=0.92\linewidth]{\footnotesize{#1}};}
       }%
       \ifthenelse{\isempty{#1}}{\vspace{2pt}}{\vspace{-2pt}}%
}

\usepackage[T1]{fontenc}

\usepackage[utf8]{inputenc}

\usepackage{microtype}

\usepackage{inconsolata}

\title{Beyond Agreement: Diagnosing the Rationale Alignment of Automated Essay Scoring Methods based on Linguistically-informed Counterfactuals}

\author{
Yupei Wang$^\dag$ \qquad 
Renfen Hu$^\dag$\thanks{\ \ Corresponding author.}{} \qquad 
Zhe Zhao$^\ddag$ \\
$\dag$ Beijing Normal University \\ 
$\ddag$ Tencent AI Lab \\
  \texttt{\{wangyupei,irishu\}@mail.bnu.edu.cn},
  \texttt{nlpzhezhao@tencent.com}
}

\begin{document}
\maketitle
\begin{abstract}
While current Automated Essay Scoring (AES) methods demonstrate high scoring agreement with human raters, their decision-making mechanisms are not fully understood.
Our proposed method, using counterfactual intervention assisted by Large Language Models (LLMs), reveals that BERT-like models primarily focus on sentence-level features, whereas LLMs such as GPT-3.5, GPT-4 and Llama-3 are sensitive to conventions \& accuracy, language complexity, and organization, indicating a more comprehensive rationale alignment with scoring rubrics.
Moreover, LLMs can discern counterfactual interventions when giving feedback on essays. 
Our approach improves understanding of neural AES methods and can also apply to other domains seeking transparency in model-driven decisions\footnote{We release our code at~\url{https://github.com/YpLarryWang/beyond-agreement-aes-2024}.}.

\end{abstract}

\section{Introduction}
\label{sec:introduction}

In recent years, neural approaches to Automated Essay Scoring (AES) have demonstrated remarkable performance~\citep{ke_automatedessayscoring_2019,ramesh_automatedessayscoring_2022}. The advent of Large Language Models (LLMs) has shifted focus not only towards their scoring capabilities but also towards the potential for providing feedback ~\citep{mizumoto_exploringpotentialusing_2023,caines_applicationlargelanguage_2023, han_fabricautomatedscoring_2023,xiao_automationaugmentationlarge_2024}, enabling a better understanding of the models' rationale. 
However, current model evaluations mainly use metrics such as Quadratic Weighted Kappa (QWK) to measure agreement with human ratings.
This approach leaves the models' underlying reasoning opaque, thereby raising risks and questioning the validity of their use in high-stakes educational tests~\citep{fiacco_extractingunderstandingimplicit_2023}.

A series of studies have found that neural models can be \emph{right for the wrong reasons}, a concern that persists into the era of LLMs~\citep{mccoy2020right, turpin2023language}. 
To understand the decision-making basis of neural models, researchers have primarily adopted two primary avenues: \textcolor{cdel}{\it what} knowledge a model encodes and \textcolor{cdel}{\it why} a model makes certain predictions~\citep{lyu2024towards}. Both paradigms have garnered attention in the field of AES.
\citet{fiacco_extractingunderstandingimplicit_2023} addresses the \textcolor{cdel}{\it what} question by extracting meaningful functional groups from the representations of transformer models and aligning them with human-understandable features. However, a model encodes a myriad of features does not mean that the features are utilized in decision-making~\citep{lyu2024towards}. 
To tackle the \textcolor{cdel}{\it why} question, \citet{singla2023automatic} employed integrated gradients~\citep{sundararajan_axiomaticattributiondeep_2017} to analyze token importance, and discovered that for BERT-based model, most of the attributions are over non-linguistic tokens and stop words. It can be seen that the gradients-based methods only target lower-level token features, thus failing to reveal whether models leverage higher-level linguistic features. 
Moreover, both \citet{ kabra_evaluationtoolkitrobustness_2022} and \citet{singla2023automatic} employed adversarial modifications to assess models, but these interventions did not target the linguistic features critical to the AES task, and they did not control for other variables that could affect essay scores during modification.
Therefore, even for traditional AES models, reliable explanations of their inner workings remain elusive. Additionally, the explainability of scoring in LLMs is largely unexplored, indicating considerable work is needed to advance our understanding of model reasoning within this domain.

\begin{figure*}[ht!]
    \centering
    \includegraphics[width=0.97\linewidth]{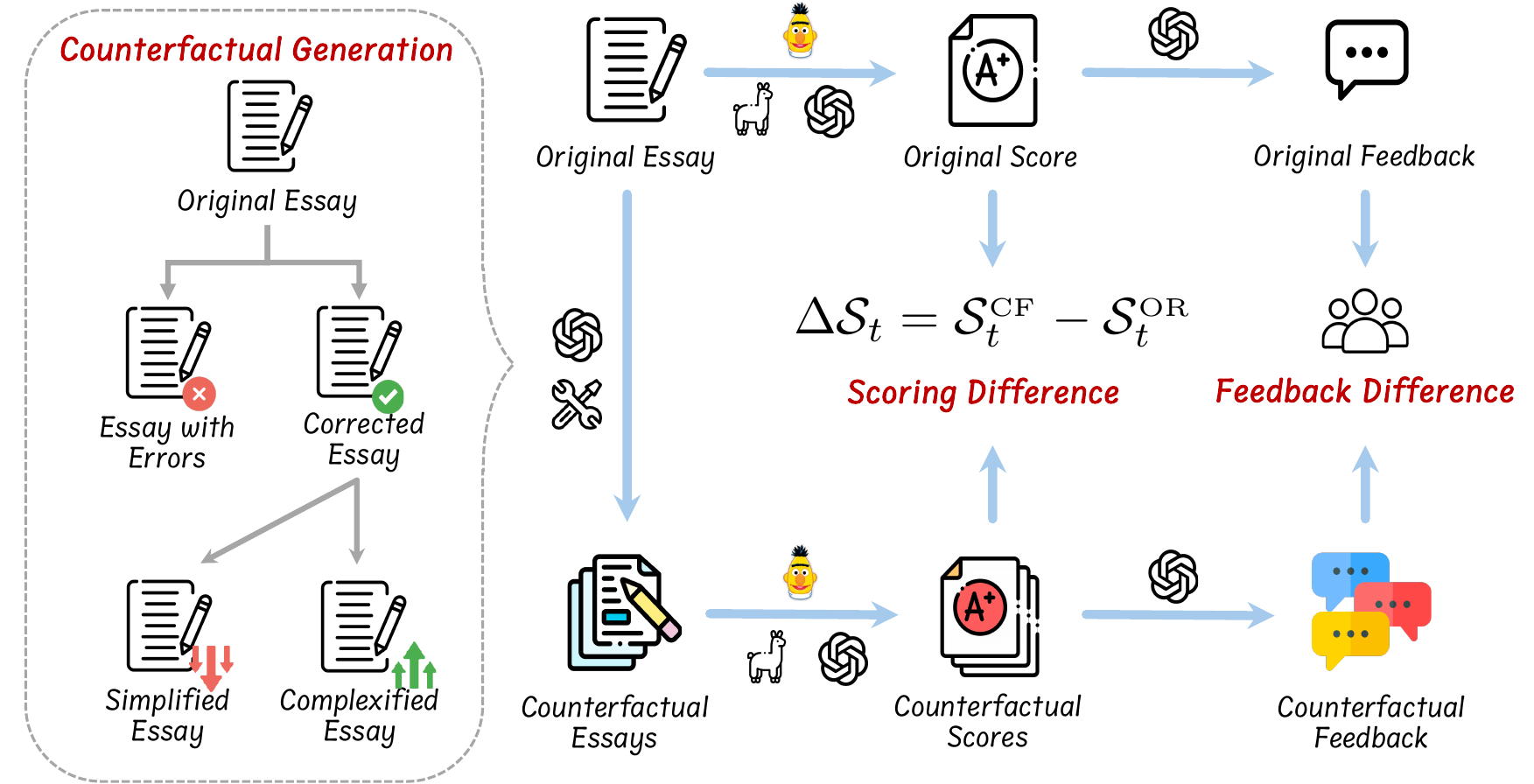}
    \caption{The pipeline of our proposed method.}
    \label{fig:overall_diagram}
    \vspace{-0.5cm}
\end{figure*}

In this paper, we systematically investigate whether models adhere to scoring rubrics when producing scores and feedback on essays, thereby assessing their alignment with human rationale. To quantify this alignment, we introduce the concept of \textit{rationale alignment}, measured by the difference in scores and feedback between original essays and their counterfactuals.
As shown in \Cref{fig:overall_diagram}, we propose a model-agnostic diagnosis method that uses linguistically-informed counterfactuals to scrutinize the scoring behavior of BERT-like models and LLMs.
The diagnostic approach closely integrates linguistic knowledge from scoring rubrics, such as conventions, accuracy, vocabulary, syntax, and coherence, with LLMs employed for fine-grained and controllable counterfactual generation.

Our investigation reveals that: \textbf{(1)} BERT-like models
can discern differences in conventions and language complexity but struggle to grasp the logical structure and coherence of essays; and
\textbf{(2)} LLMs, although have lower score agreement than traditional models, display a superior alignment with human experts' reasoning during scoring and can also address counterfactual interventions in their feedback.
Through few-shot learning or fine-tuning, LLMs can achieve both high scoring agreement and rationale alignment.

\section{Related Work}
\label{sec:related_work}

\subsection{AES based on Neural Language Models}
\label{subsec:higher_agreement}

Pre-trained neural language models have made significant progress in the field of AES. After fine-tuning on specific datasets, these models can achieve high levels of agreement with human raters~\citep{rodriguez_languagemodelsautomated_2019,yang_enhancingautomatedessay_2020,ormerod_automatedessayscoring_2021,wang_usebertautomated_2022}.
Since the emergence of ChatGPT, the scoring performance of LLMs has garnered considerable attention. 
Leveraging their powerful language understanding capabilities and in-context learning abilities, LLMs can evaluate essays and assign overall scores or scores for specific dimensions~\citep{naismith_automatedevaluationwritten_2023}. 
However, research has shown that zero-shot and few-shot LLMs fail to achieve state-of-the-art scoring performance~\citep{mizumoto_exploringpotentialusing_2023}, while fine-tuned LLM models exhibit notable superiority~\citep{xiao_automationaugmentationlarge_2024}.

Although the scoring ability of LLMs without fine-tuning is not particularly remarkable, they can provide explainable feedback in natural language. 
Previously, essay feedback was primarily provided through trait scores (e.g., vocabulary) \citep{carlile_givememore_2018,hussein_traitbaseddeeplearning_2020,lee_humancentricautomatedessay_2023}. With the emergence of LLMs, researchers discovered that it is possible to elicit explanations about assessment decisions from the models~\citep{caines_applicationlargelanguage_2023}. \citet{han_fabricautomatedscoring_2023} assessed the feedback generated by GPT-3.5 on level of detail, accuracy, relevance, and helpfulness, while \citet{xiao_automationaugmentationlarge_2024} found that GPT-4 feedback could elevate novice raters to expert levels.

\label{subsec:feedback_provision}

\subsection{Interpretability and Robustness of AES Models}
\label{subsec:interpretability}

In terms of model interpretability in AES research, \citet{fiacco_extractingunderstandingimplicit_2023} analyzed the features encoded by transformer models, but this approach provides limited insight into the decision-making rationale of the models. \citet{singla2023automatic} employed the integrated gradients~(IG) method~\citep{sundararajan_axiomaticattributiondeep_2017} on neural models to analyze token-level feature importance and discovered that BERT-based models frequently assign substantial importance to stopwords and non-linguistic tokens. This counterintuitive result may stem from the fact that the IG method does not address interactions between tokens, thereby failing to capture abstract linguistic concepts such as cohesion and syntax. Moreover, these methods cannot be directly applied to closed-source models like GPT-3.5 and GPT-4.

Additionally, a line of works have utilized adversarial modifications to diagnose model robustness. \citet{powers2002stumping} invited human writers to compose essays that would ''trick'' the AES system and found that repeating, rewording, and reordering were effective strategies. \citet{bejar2014vulnerability} employed the substitution of words with less frequent and longer synonyms. \citet{kabra_evaluationtoolkitrobustness_2022} used methods such as the addition of irrelevant lines, the introduction of grammatical errors, and the deletion of lines from the responses. \citet{myers2023evaluating} evaluated models using a sentence-level randomization approach. It is important to note that these studies aim to expose model vulnerabilities by introducing input perturbations rather than exploring the interpretability of model decisions.

\subsection{Counterfactual Analysis}
Counterfactuals are hypothetical scenarios created to understand the causal effects of specific interventions in a given situation~\citep{feder_causalinferencenatural_2022}. Existing counterfactual generation methods utilize keyword replacement \citep{garg2019counterfactual}, sentence rewriting \citep{ross2021tailor, wu2021polyjuice}, and manual editing \citep{gardner2020evaluating}. However, these approaches are often limited to simple local interventions or require costly manual annotation, which hinders the practical estimation of the causal effects of high-level concepts on NLP models. While recent efforts have leveraged LLMs for generating more natural and diverse counterfactuals~\citep{dixit2022core,chen2023disco}, most have only exploited LLMs' powerful language generation capabilities without tapping into their potential to understand and manipulate abstract concepts within texts.
\citet{gat_faithfulexplanationsblackbox_2023} found that LLMs can produce high-quality counterfactuals, which assist in providing strong black-box model explanations. \citet{li2024prompting} prompted LLMs to identify and modify \textit{causal terms} to generate counterfactuals. Inspired by these works, we decided to combine LLMs with rule-based methods to achieve controlled sample generation in AES.

\vspace{-0.1cm}
\section{Method}
\vspace{-0.1cm}
\label{sec:method}

\begin{table*}[!t]\small
    \centering
    \setlength{\abovecaptionskip}{0.2cm}
    \resizebox{\linewidth}{!}{\begin{tabular}{lll} 
\hline
\textbf{Concept}         & \textbf{Intervention} & \textbf{Description} \\ \hline
\multirow{4}{2cm}{Conventions \\\& Accuracy}      & \textcolor{cadd}{Error Correction}  & Prompt GPT-4 Turbo to correct spelling, punctuation, and grammar errors.\\       
    & \textcolor{cdel}{Spelling Errors Introduction}   & Use \texttt{nlpaug}~\citep{ma2019nlpaug} to misspell 30\% of words in 50\% of sentences.\\ 
    & \textcolor{cdel}{Agreement Errors Introduction} & Introduce subject-verb agreement (SVA) errors in 50\% of sentences.  \\
    & \textcolor{cdel}{Word Order Swapping (WOS)} & Use \texttt{nlpaug} to swap 30\% of words in 50\% of sentences. \\
\hline
\multirow{2}{2cm}{Language Complexity} & \textcolor{cadd}{Complexification}   & Prompt GPT-4 Turbo to enhance vocabulary and sentence structure.                         \\
   & \textcolor{cdel}{Simplification}   & Prompt GPT-4 Turbo to simplify vocabulary and sentence structure. \\
\hline
\multirow{2}{2cm}{Organization \\\& Development}   & \textcolor{cdel}{Intra-paragraph Shuffling}   & Shuffle sentence order \textit{within paragraphs} to disrupt \textit{local} cohesion.                                      \\
   & \textcolor{cdel}{Inter-text Shuffling} & Shuffle sentence order \textit{across the entire essay} to disrupt \textit{global} cohesion. \\ \hline
\end{tabular}
}
    \caption{Overview of \textcolor{cadd}{positive} and \textcolor{cdel}{negative} counterfactual intervention methods used.}
\label{tab:generation_methods}
\vspace{-0.5cm}
\end{table*}

We employed counterfactual interventions to establish causality between target concepts and predicted scores. Typically, counterfactual intervention involves manipulating a specific feature or concept while controlling for others and observing the subsequent effect on the model's prediction. 
We firstly extracted target concepts from scoring rubrics for intervention, and then generated counterfactual samples for different concepts using LLMs and heuristic rules.

\vspace{-0.1cm}
\subsection{Concepts for Intervention}

To identify the target concepts for AES scenarios, we reviewed scoring rubrics from major standardized English tests (IELTS, TOEFL iBT, TOEIC, PTE Academic) and the ELLIPSE dataset~\cite{crossley2023ellipse}, which is based on various state and industrial English language proficiency assessments. 
We conducted a detailed annotation process to identify common linguistic features across the five rubrics. See Appendix~\ref{app:rubric} for more information. Through this analysis, we discovered that all the scoring criteria consistently emphasize three key aspects:

\textbf{Conventions and Accuracy}: An essay is considered to adhere to conventions and demonstrate accuracy when it is free from mechanical (spelling, capitalization, and punctuation) mistakes and grammatical inaccuracies.

\textbf{Language Complexity}: An essay demonstrates lexical and syntactic complexity through the use of a broad vocabulary, sophisticated lexical control, and varied sentence structures.

\textbf{Organization and Development}: An essay exhibits effective organization and development by presenting a logical structure with skillful paragraphing and the use of cohesive devices to ensure unity, progression, and seamless connection of thoughts.

\subsection{Measurement of Rationale Alignment}

We introduce two methods—\textit{score differences} and \textit{feedback differences}—to measure rationale alignment, assessing how well models adhere to human-defined scoring rubrics.

\textbf{Score Differences}: For each concept $ C_i $, let $ O_i = \{ o_1^{(i)}, o_2^{(i)}, \ldots, o_k^{(i)} \} $ be the set of possible types of interventions. Each $ o_j^{(i)} $ represents a specific way to alter the value $ v_i $ of concept $ C_i $ in essay $ t $, resulting in a modified value $ v_i^{\prime} = o_j^{(i)}(v_i) $.

For each essay $ t $ in the test set $ \mathcal{T} $, we generate multiple counterfactual versions by altering the values of different concepts $ \{ C_i \} $ using various interventions $ \{ o_j^{(i)} \} $. We use a model $ \mathcal{M} $ to predict scores for both the original essay and its counterfactuals. The effect of a specific counterfactual intervention on model $ \mathcal{M} $ with respect to essay $ t $ is calculated by subtracting the original score from the counterfactual score:
\begin{equation}
    \Delta \mathcal{S}_t^{\mathcal{M}}\left(C_i, o_j^{(i)}\right) = \mathcal{S}_t^{\mathcal{M}}\left(C_i = v_i^{\prime}\right) - \mathcal{S}_t^{\mathcal{M}}\left(C_i = v_i\right)
    \label{eq:cf_effect}
\end{equation}
We then compute the mean effect of this specific intervention on model $ \mathcal{M} $ across all essays in $ \mathcal{T} $:
\begin{equation}
    \Delta \mathcal{S}_{\mathcal{T}}^{\mathcal{M}}\left(C_i, o_j^{(i)}\right) = \frac{1}{|\mathcal{T}|} \sum_{t \in \mathcal{T}} \Delta \mathcal{S}_t^{\mathcal{M}}\left(C_i, o_j^{(i)}\right)
\end{equation}

\textbf{Feedback Differences}: We manually compare the feedback provided for both the original and counterfactual essays. This helps us understand shifts in the model's reasoning and justification.

\subsection{Counterfactual Generation}

We employ a hybrid approach combining rule-based and LLM-based methods to generate eight types of linguistically informed counterfactuals for diagnostic purposes, as detailed in \Cref{tab:generation_methods}.
These interventions derive from three aforementioned linguistic concepts and are implemented in both positive and negative directions for conventions and language complexity. 
As shown in Figure~\ref{tab:generation_methods}, for conventions and accuracy, we introduce errors such as spelling, subject-verb agreement, and word order for negative impacts, and use LLMs to correct all errors for positive impacts.
Regarding the language complexity, we leverage LLMs to increase and decrease the language complexity along both vocabulary and syntax dimensions, building upon the basis of error correction..
For the organizational aspect, negative interventions include disrupting the sentence order within paragraphs to affect local coherence and across the entire article to impact global coherence. See \Cref{app:subsec:cfact_gen_prompt} for LLM prompts used to generate counterfactuals.

\subsection{The Validity of LLM Generated Counterfactuals}
\label{subsec:validity_cf_gen}

As shown in \Cref{tab:generation_methods}, we prompted LLMs to correct errors, complexify, and simplify essays to manipulate their conventions and language complexity.
To evaluate counterfactual essays generated by LLMs, we introduced seven linguistic metrics that measure the essay length, lexical diversity, lexical sophistication, syntactic complexity and writing error density. The descriptions of these metrics can be seen in \Cref{app:tab:lingfeat}.
We compute Cohen’s $\mathcal{D}$~\citep{cohen2013statistical} effect size for each metric as follows:
\begin{equation}
    \mathcal{D}=\frac{\bar{x}_{_{CF}}-\bar{x}_{_{OR}}}{s}
\end{equation}
where $\bar{x}_{_{CF}}$ and $\bar{x}_{_{OR}}$ are the mean values of a metric for the counterfactual and original samples, and the pooled standard deviation $s$ is defined as:
\begin{equation}
s=\sqrt{\frac{\left(n_{_{_{OR}}}-1\right) s_{_{OR}}^2+\left(n_{_{CF}}-1\right) s_{_{CF}}^2}{n_{_{OR}}+n_{_{CF}}-2}}
\end{equation}
where $ n_{_{OR}} $ and $ n_{_{CF}} $ are the sample sizes, and $ s_{_{OR}}^2 $ and $ s_{_{CF}}^2 $ are the variances of the original and counterfactual samples respectively.

\begin{table*}[t]
\centering
    \resizebox{\linewidth}{!}{\begin{tabular}{>{\footnotesize\sffamily\centering\arraybackslash}m{2.25cm} >{\small\raggedright\arraybackslash}m{14cm}} 
\toprule
\text{\bf \small Metric}          & \textbf{Description} \\
\midrule
WordNum & The number of words in an essay. \\
\hline
SentNum & The number of sentences in an essay. \\
\hline
MLS & Mean length of sentences. The length of each sentence is the number of words it has. \\
\hline
ADDT & Average depth of dependency tree for all sentences in an essay. \\
\hline
LemmaTTR & A \textit{lexical diversity} measure based on the Type-Token Ratio (TTR) of an essay, where each word is lemmatized.\\
\hline
LexSoph & A \textit{lexical sophistication} measure based on word frequency statistics from the 1980s-2010s COHA corpus~\cite{davies2010coha}. For an essay with $N$ words, let $w_1, w_2, \ldots, w_N$ be the individual words (including repetitions), $\ell_i$ be the lemma of $w_i$, and $\text{Freq}(\ell_i)$ be the frequency of $\ell_i$ in the selected COHA subset. \textsf{LexSoph} is defined as:
\[
\frac{1}{N} \sum_{i=1}^{N} \frac{1}{\log(\text{Freq}(\ell_i) + 1)}
\]
\\
\hline
ErrorDensity & Density of writing errors in an essay with $N$ words, defined as ${\#{\text{error}}} / N$. 
Writing error analyses are implemented using \texttt{LanguageTool}~\citep{naber2003languagetool}.
\\
\bottomrule
\end{tabular}

}
    \setlength{\abovecaptionskip}{0.1cm}
    \caption{The linguistics metrics used for the evaluation of counterfactual samples.}
    \label{app:tab:lingfeat}
    \vspace{-0.2cm}
\end{table*}

Moreover, we assess content preservation during interventions by calculating the average cosine similarity between text embeddings of "original-counterfactual" essay pairs for each of the three LLM-based interventions.

\begin{figure*}[ht!]
    \centering
    \setlength{\abovecaptionskip}{0.cm} 
    \includegraphics[width=\linewidth]{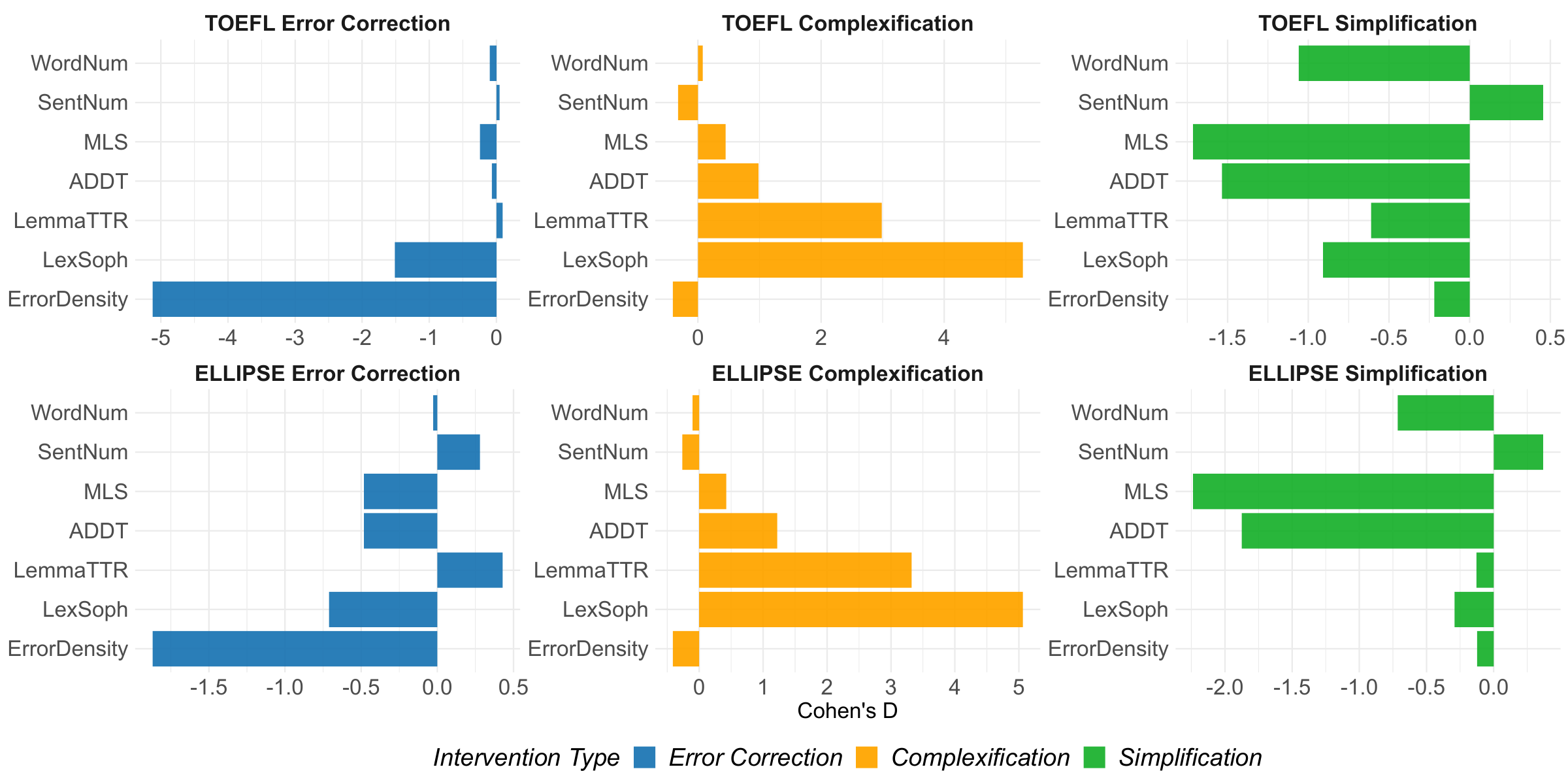}
    \caption{Cohen's $\mathcal{D}$ measured for seven linguistic metrics on three interventions.}
    \label{fig:cf_cohensd}
    \vspace{-0.5cm}
\end{figure*}

\vspace{-0.1cm}
\section{Experiments}
\label{sec:experiments}
\vspace{-0.1cm}
\subsection{Settings}
\label{subsec:settings}

Our study utilized TOEFL11~\citep{blanchard2013toefl11} and ELLIPSE datasets.
TOEFL11 includes 12,100 essays from the 2006-2007 TOEFL exams, divided into 9,900 for training, 1,100 for validation, and 1,100 for testing, with essays categorized into low, medium, or high proficiency by human raters. We assessed performance using weighted F1 and quadratic weighted kappa (QWK).
The ELLIPSE dataset contains 6,482 essays from 8th to 12th-grade English learners, with 2,568 reserved for testing. Essays were rated on a 1 to 5 scale (with 0.5 increments), adjusted to the nearest 0.5 for QWK calculations, alongside Root Mean Square Error (RMSE) evaluation.

The counterfactuals were generated on the test set using GPT-4 Turbo and Llama-3-70b-Instruct models. Comparative analysis revealed that both models successfully completed the task, but the GPT-4 Turbo model exhibited more stable performance in the aforementioned measures across both datasets (see \Cref{app:subsec:compare_gpt4_llama_cf_gen} for detailed comparisons). Consequently, we employed the counterfactual essays generated by the GPT-4 Turbo model for subsequent analyses.

For automated scoring, we fine-tuned BERT~\citep{devlin_bertpretrainingdeep_2019}, RoBERTa~\cite{liu_robertarobustlyoptimized_2019} and DeBERTa~\citep{he_debertadecodingenhancedbert_2021} on the training set. 
For LLMs, we utilized GPT-3.5 Turbo, GPT-4 Turbo and Llama 3 instruction-fine-tuned models (8B \& 70B) in zero-shot learning (ZSL) and few-shot learning (FSL) scenarios, and performed supervised fine-tuning (SFT) on GPT-3.5 Turbo. 
Detailed fine-tuning and inference settings are provided in \Cref{app:sec:scoring_implement_detail}.

Since score levels of TOEFL11 essays are labeled as three discrete categories\footnote{When few-shot prompting LLMs to rate TOEFL11 essays, score levels of example essays in prompts can only be one of the three categories. Therefore it is not possible to expect numerical predicted scores, which is the reason why in \Cref{tab:overall_cfact_results} we do not show intervention effects of FSL on TOEFL11.}: low, medium and high, we calculated the score delta of BERT-like models after converting the classifier's output probabilities into 1-5 scores for consistency with TOEFL scoring rubrics scale:
\begin{equation}
\mathcal{S}^{\mathcal{M}}_{t}=\sum_{k=1}^3(2 k-1) \operatorname{Pr}^{\mathcal{M}}(\hat{y}_{t}=k)
\end{equation}
where $\hat{y}_{t}$ is the predicted class of essay $t$.

\subsection{Counterfactual Validation Results}
\label{subsec:results_of_cfact_gen_valid}

\Cref{fig:cf_cohensd} shows the effect size of three GPT-4-based interventions on seven linguistic metrics across two datasets.
Both datasets show similar patterns, with Error Correction notably reducing error density and leaving lexical and syntactic complexity almost untouched.
Complexification significantly boosts lexical diversity and sophistication while moderately enhancing syntactic complexity, without substantially changing overall text length and error density.
Similarly, Simplification effectively reduces sentence length and complexity while also making corresponding changes to lexical properties.

\begin{table*}[ht]
\centering
\setlength{\abovecaptionskip}{0.1cm}
\scriptsize
\resizebox{\linewidth}{!}{
\begin{tabular}{>{\sffamily\raggedright\arraybackslash}m{15cm}}
\toprule
\textbf{Original:}\quad
In the \textcolor{cdel}{up coming} years \textcolor{cdel}{i} \textcolor{cdel}{wanna} save up and work hard to \textcolor{cdel}{get} money, so that \textcolor{cdel}{i} can open my own personal store.\textcolor{cdel}{ .}
\textbackslash n\textbackslash n
\textcolor{cdel}{in} the store \textcolor{cdel}{i} \textcolor{cdel}{going} to \textcolor{cdel}{be} selling high \textcolor{cdel}{branded} shoes\textcolor{cdel}{,}clothes\textcolor{cdel}{,} and \textcolor{cdel}{outer collectives}. \textcolor{cdel}{and} \textcolor{cdel}{Im} going to \textcolor{cdel}{learning} how to invest my money as \textcolor{cdel}{i} sell things before opening up my dream store.
\textbackslash n\textbackslash n
\textcolor{cdel}{i wanna} do this as a job because \textcolor{cdel}{ive awlays had a passhine} for shoes and high \textcolor{cdel}{branded} things. \textcolor{cdel}{and if i} can pull this off \textcolor{cdel}{i think ill be veary succes full} in life \textcolor{cdel}{i will be getting and veary good salary ill be able to pay off anythig and have to money} to support my family.
 \\
\midrule
\textbf{Error Correction:}\quad
"In the \textcolor{cadd}{upcoming} years\textcolor{cadd}{,} \textcolor{cadd}{I} \textcolor{cadd}{want to} save up and work hard to \textcolor{cadd}{earn} money\textcolor{cadd}{,} so that \textcolor{cadd}{I} can open my own personal store.
\textbackslash n\textbackslash n
\textcolor{cadd}{In} the store\textcolor{cadd}{,} \textcolor{cadd}{I am going to sell} high\textcolor{cadd}{-branded} shoes\textcolor{cadd}{,} clothes\textcolor{cadd}{, and other collectibles.} \textcolor{cadd}{I'm also} going to \textcolor{cadd}{learn} how to invest my money as \textcolor{cadd}{I} sell things before opening up my dream store.
\textbackslash n\textbackslash n
\textcolor{cadd}{I want to} do this as a job because \textcolor{cadd}{I've always had a passion} for shoes and high\textcolor{cadd}{-branded} \textcolor{cadd}{items}. \textcolor{cadd}{If I} can pull this off\textcolor{cadd}{, I think I'll be very successful} in life\textcolor{cadd}{. I will be earning a very good salary, I'll be able to pay off anything, and have the money} to support my family."
 \\
\midrule
\textbf{Complexification:}\quad
In the forthcoming years, my ambition is to accumulate sufficient capital and diligently strive to generate income, thereby enabling me to inaugurate my own boutique. Within this emporium, I intend to offer an array of premium-branded footwear, apparel, and assorted collectibles. Concurrently, I shall endeavor to acquire financial acumen, particularly in investment strategies, as I engage in commerce prior to the launch of my envisioned enterprise. Pursuing this career path is fueled by an enduring zeal for exclusive footwear and luxury goods. Should I succeed in actualizing this venture, I am confident that it will herald substantial success in my life. Not only will I secure a lucrative income, but I will also possess the financial capacity to discharge any obligations and provide for my family's needs.
 \\
\midrule
\textbf{Simplification:}\quad
In the next few years, I plan to save money and work hard. I want to make enough money to open my own store. In my store, I will sell popular shoes, clothes, and other special items. I will also learn how to use my money well as I sell things. I want to open my dream store one day. I have always loved shoes and popular brands. If I can do this, I think I will do very well in life. I will make a lot of money. I will be able to pay for everything and have enough to take care of my family.
 \\
\bottomrule
\end{tabular}
}
\caption{Example of a 3-point ELLIPSE essay and its counterfactual counterparts generated by GPT-4 Turbo.}
\label{tab:cfact_example}
\vspace{-0.5cm}
\end{table*}

\Cref{tab:cos_sim_gpt4_cf} presents the embedding similarities bewteen counterfactuals and original essays.
It can be seen that Error Correction almost completely retains the original meaning, and Complexification and Simplification, although inevitably making more changes to the original text, still retain most of the original meaning.
For better clarification, \Cref{tab:cfact_example} shows counterfactual examples of a \textit{3-point} ELLIPSE essay generated by GPT-4 Turbo. Examples of its rule-based counterfactuals can be seen in \Cref{app:subsec:rule_based_cf_examples}.

\begin{table}[h!]
    \centering
    \setlength{\abovecaptionskip}{0.1cm}
    \resizebox{0.75\columnwidth}{!}{\begin{tabular}[t]{lcc}
\toprule
\textbf{Intervention} & \textbf{TOEFL11} & \textbf{ELLIPSE} \\
\midrule
Error Correction & 0.935 & 0.942 \\
Complexification & 0.760 & 0.749 \\
Simplification & 0.816 & 0.849 \\
\bottomrule
\end{tabular}
}
    \caption{Content preservation for GPT-4-based interventions: text cosine similarities computed by OpenAI \texttt{text-embedding-3-large}.}
    \label{tab:cos_sim_gpt4_cf}
    \vspace{-0.6cm}
\end{table}

\subsection{Scoring Results}
\label{subsec:agreement_and_alignment}

\Cref{tab:model_performance} displays the performance of scoring agreement between models and human on the test sets of both datasets. Note that the \textsc{GPT-3.5-sft-100} setting is fine-tuned on 100 essays randomly selected from the overall training set, ensuring a stratified distribution based on essay scores. \Cref{tab:overall_cfact_results} shows intervention effects of different types of counterfactual interventions. Based on these results, our findings are as follows:

\begin{table}[h!]
    \centering
    \setlength{\abovecaptionskip}{0.1cm}
    \resizebox{0.98\columnwidth}{!}{\begin{tabular}[t]{lcccc}
\toprule
 \multirow[c]{2}{*}{\bf \large Setting} & \multicolumn{2}{c}{\textbf{TOEFL11}} & \multicolumn{2}{c}{\textbf{ELLIPSE}} \\
 \cmidrule(r){2-3} \cmidrule(r){4-5}
 &  \textbf{F1} $\uparrow$ & \textbf{QWK} $\uparrow$ & \textbf{RMSE} $\downarrow$ & \textbf{QWK} $\uparrow$ \\
\midrule
\midrule
\textsc{BERT} & 0.783 & 0.736 & 0.437 & 0.680 \\
\textsc{RoBERTa} & \textbf{0.795} & 0.739 & 0.430 & 0.695 \\
\textsc{DeBERTa} & 0.790 & \textbf{0.741} & \textbf{0.422} & \textbf{0.720} \\
\midrule
\textsc{GPT-3.5-zsl} & 0.599 & 0.408 & 0.701 & 0.399 \\
\textsc{GPT-3.5-fsl} & 0.546 & 0.314 & \underline{0.570} & 0.378 \\
\textsc{GPT-3.5-sft-100} & \colorbox{green!15}{0.710} & \colorbox{green!15}{0.592} & \colorbox{green!15}{0.550} & \colorbox{green!15}{0.629} \\
\textsc{GPT-4-zsl} & 0.368 & 0.380 & 0.960 & 0.261 \\
\textsc{GPT-4-fsl} & 0.490 & 0.477 & 0.680 & 0.466 \\
\textsc{Llama-3-8b-zsl} & 0.558 & 0.297 & 0.628 & 0.345 \\
\textsc{Llama-3-8b-fsl} & 0.435 & 0.441 & 1.039 & 0.054 \\
\textsc{Llama-3-70b-zsl} & 0.524 & 0.390 & 0.903 & 0.182 \\
\textsc{Llama-3-70b-fsl} & \underline{0.609} & \underline{0.562} & 0.589 & \underline{0.503} \\
\bottomrule
\end{tabular}
}
    \caption{The scoring agreement performance on both test sets: \textbf{best setting} in bold, \colorbox{green!15}{fine-tuned GPT-3.5} with a green shadow, \underline{best off-the-shelf LLMs} underlined.
    Metrics with $\uparrow$ indicate that higher values are better, while the one with $\downarrow$ indicates that lower values are better.}
    \label{tab:model_performance}
    \vspace{-0.3cm}
\end{table}

\begin{table*}[h!]
    \centering
    \setlength{\abovecaptionskip}{0.1cm}
    \resizebox{\linewidth}{!}{\begin{tabular}[t]{llcccccccc}
\toprule
& & \multicolumn{4}{c}{\textbf{Conventions \& Accuracy}} & \multicolumn{2}{c}{\textbf{Language Complexity}} & \multicolumn{2}{c}{\textbf{Organization \& Development}} \\
\cmidrule(r){3-6} \cmidrule(r){7-8} \cmidrule(r){9-10}
\textbf{Dataset} & \textbf{Setting} & \textbf{Error Correction~$(+)$} & \multicolumn{3}{c}{\textbf{Error Introduction~$(-)$}} & \textbf{Complexification~$(+)$} & \textbf{Simplification~$(-)$} & \textbf{InParaShuffle~$(-)$} & \textbf{InTextShuffle~$(-)$}\\
\cmidrule(r){4-6}
& & -- & \textbf{Spelling} & \textbf{SVA} & \textbf{WOS} & -- &-- &-- &-- \\
\midrule
\midrule
\multirow[c]{8}{*}{TOEFL11}
 & \textsc{BERT} & $1.03_{-.041}^{+.043}$ & $-0.92_{-.033}^{+.032}$ & $-0.22_{-.014}^{+.013}$ & $-1.26_{-.032}^{+.033}$ & $0.42_{-.035}^{+.035}$ & $-0.69_{-.033}^{+.033}$ & $-0.01_{-.006}^{+.006}$ & $-0.01_{-.006}^{+.006}$ \\
 & \textsc{RoBERTa} & $0.99_{-.044}^{+.043}$ & $-0.79_{-.032}^{+.033}$ & $-0.45_{-.021}^{+.021}$ & $-1.13_{-.033}^{+.033}$ & $0.24_{-.031}^{+.032}$ & $-0.35_{-.025}^{+.025}$ & $-0.19_{-.011}^{+.010}$ & $-0.02_{-.005}^{+.005}$ \\
 & \textsc{DeBERTa} & $1.19_{-.046}^{+.045}$ & $-0.92_{-.031}^{+.031}$ & $-0.35_{-.016}^{+.016}$ & $-1.24_{-.032}^{+.033}$ & $0.33_{-.032}^{+.034}$ & $-0.27_{-.026}^{+.027}$ & $-0.06_{-.005}^{+.005}$ & $-0.06_{-.005}^{+.005}$ \\
 \cmidrule{2-10}
 & \textsc{GPT-3.5-zsl} & $0.64_{-.031}^{+.032}$ & $-0.76_{-.034}^{+.033}$ & $-0.20_{-.026}^{+.026}$ & $-0.59_{-.030}^{+.032}$ & $0.27_{-.024}^{+.025}$ & {\color{black!30} $0.01_{-.020}^{+.019}$ } & $-0.31_{-.030}^{+.030}$ & $-0.42_{-.032}^{+.032}$ \\
 & \textsc{GPT-4-zsl} & $0.92_{-.025}^{+.025}$ & $-0.80_{-.025}^{+.025}$ & $-0.35_{-.021}^{+.021}$ & $-0.80_{-.026}^{+.026}$ & $0.66_{-.025}^{+.025}$ & $-0.24_{-.021}^{+.021}$ & $-0.24_{-.017}^{+.018}$ & $-0.29_{-.019}^{+.019}$ \\
 & \textsc{Llama-3-8b-zsl} & $0.58_{-.026}^{+.027}$ & $-0.37_{-.029}^{+.029}$ & $-0.07_{-.018}^{+.018}$ & $-0.17_{-.024}^{+.023}$ & $0.57_{-.026}^{+.026}$ & $-0.11_{-.023}^{+.023}$ & $-0.15_{-.024}^{+.024}$ & $-0.23_{-.026}^{+.026}$ \\
 & \textsc{Llama-3-70b-zsl} & $0.64_{-.025}^{+.026}$ & $-0.56_{-.025}^{+.025}$ & $-0.24_{-.022}^{+.021}$ & $-0.41_{-.023}^{+.023}$ & $1.19_{-.032}^{+.032}$ & $-0.17_{-.024}^{+.024}$ & $-0.15_{-.019}^{+.019}$ & $-0.19_{-.021}^{+.021}$ \\
 \cmidrule{1-10}
\multirow[c]{16}{*}{ELLIPSE} 
 & \textsc{BERT} & $0.84_{-.014}^{+.014}$ & $-0.57_{-.011}^{+.011}$ & $-0.09_{-.003}^{+.003}$ & $-0.57_{-.011}^{+.011}$ & $0.31_{-.009}^{+.009}$ & $-0.11_{-.008}^{+.008}$ & $-0.01_{-.002}^{+.002}$ & $-0.02_{-.003}^{+.002}$ \\
 & \textsc{RoBERTa} & $0.92_{-.015}^{+.014}$ & $-0.50_{-.009}^{+.009}$ & $-0.11_{-.003}^{+.003}$ & $-0.54_{-.009}^{+.009}$ & $0.25_{-.007}^{+.008}$ & $-0.05_{-.007}^{+.007}$ & $-0.01_{-.002}^{+.002}$ & $-0.10_{-.003}^{+.003}$ \\
 & \textsc{DeBERTa} & $1.06_{-.016}^{+.016}$ & $-0.64_{-.013}^{+.013}$ & $-0.20_{-.006}^{+.006}$ & $-0.64_{-.013}^{+.013}$ & \textcolor{cdel}{$-0.08_{-.007}^{+.007}$} & \textcolor{cdel}{$0.01_{-.005}^{+.005}$} & $-0.02_{-.001}^{+.001}$ & $-0.07_{-.002}^{+.002}$ \\
 \cmidrule{2-10}
 & \textsc{GPT-3.5-zsl} & $0.77_{-.018}^{+.019}$ & $-0.60_{-.018}^{+.019}$ & $-0.19_{-.015}^{+.015}$ & $-0.35_{-.018}^{+.018}$ & $0.48_{-.016}^{+.016}$ & \textcolor{cdel}{$0.08_{-.014}^{+.014}$} & $-0.15_{-.014}^{+.015}$ & $-0.18_{-.017}^{+.016}$ \\
 & \textsc{GPT-3.5-fsl} & $0.35_{-.014}^{+.014}$ & $-0.46_{-.015}^{+.015}$ & $-0.15_{-.012}^{+.012}$ & $-0.31_{-.014}^{+.014}$ & $0.36_{-.014}^{+.014}$ & $-0.04_{-.012}^{+.012}$ & $-0.11_{-.012}^{+.013}$ & $-0.16_{-.014}^{+.014}$ \\
 & \textsc{GPT-4-zsl*} & $0.87_{-.058}^{+.060}$ & $-0.64_{-.047}^{+.047}$ & $-0.30_{-.045}^{+.045}$ & $-0.56_{-.045}^{+.045}$ & $0.96_{-.065}^{+.065}$ & {\color{black!30} $-0.05_{-.057}^{+.058}$ } & $-0.10_{-.035}^{+.033}$ & $-0.19_{-.040}^{+.037}$ \\
 & \textsc{GPT-4-fsl*} & $0.61_{-.048}^{+.052}$ & $-0.71_{-.060}^{+.060}$ & $-0.27_{-.050}^{+.050}$ & $-0.56_{-.050}^{+.048}$ & $0.67_{-.052}^{+.055}$ & $-0.09_{-.043}^{+.045}$ & $-0.14_{-.035}^{+.032}$ & $-0.23_{-.045}^{+.042}$ \\
 & \textsc{Llama-3-8b-zsl} & $0.32_{-.016}^{+.017}$ & $-0.31_{-.018}^{+.018}$ & $-0.06_{-.011}^{+.011}$ & $-0.11_{-.014}^{+.013}$ & $0.70_{-.013}^{+.013}$ & {\color{black!30} $0.01_{-.010}^{+.009}$ } & $-0.06_{-.012}^{+.011}$ & $-0.10_{-.014}^{+.014}$ \\
 & \textsc{Llama-3-8b-fsl} & $0.06_{-.011}^{+.011}$ & $-0.11_{-.016}^{+.016}$ & $-0.02_{-.008}^{+.008}$ & $-0.06_{-.011}^{+.011}$ & $0.07_{-.016}^{+.016}$ & {\color{black!30} $-0.00_{-.007}^{+.007}$ } & $-0.02_{-.010}^{+.010}$ & $-0.02_{-.011}^{+.012}$ \\
 & \textsc{Llama-3-70b-zsl*} & $0.51_{-.018}^{+.018}$ & $-0.41_{-.011}^{+.011}$ & $-0.11_{-.009}^{+.009}$ & $-0.19_{-.010}^{+.010}$ & $1.63_{-.019}^{+.019}$ & \textcolor{cdel}{$0.03_{-.018}^{+.018}$} & $-0.03_{-.007}^{+.007}$ & $-0.06_{-.008}^{+.008}$ \\
 & \textsc{Llama-3-70b-fsl*} & $0.51_{-.068}^{+.070}$ & $-0.54_{-.065}^{+.065}$ & $-0.12_{-.035}^{+.033}$ & $-0.24_{-.052}^{+.050}$ & $1.08_{-.055}^{+.055}$ & {\color{black!30} $-0.04_{-.040}^{+.040}$ } & $-0.11_{-.042}^{+.040}$ & $-0.13_{-.045}^{+.043}$ \\
 \cmidrule{2-10}
 & \textsc{GPT-3.5-sft-50*} & $0.83_{-.072}^{+.075}$ & $-0.64_{-.080}^{+.077}$ & $-0.14_{-.050}^{+.045}$ & $-0.34_{-.068}^{+.065}$ & $0.96_{-.062}^{+.060}$ & \textcolor{cdel}{$0.08_{-.052}^{+.055}$} & $-0.09_{-.045}^{+.045}$ & $-0.10_{-.050}^{+.047}$ \\
& \textsc{GPT-3.5-sft-100*} & $1.12_{-.080}^{+.080}$ & $-0.95_{-.080}^{+.080}$ & $-0.26_{-.052}^{+.052}$ & $-0.58_{-.055}^{+.057}$ & $0.88_{-.057}^{+.055}$ & {\color{black!30} $0.05_{-.048}^{+.050}$ } & $-0.18_{-.050}^{+.047}$ & $-0.19_{-.050}^{+.048}$ \\
& \textsc{GPT-3.5-sft-200*} & $1.03_{-.090}^{+.092}$ & $-0.57_{-.090}^{+.087}$ & {\color{black!30} $-0.01_{-.070}^{+.068}$ } & $-0.32_{-.070}^{+.072}$ & $0.79_{-.055}^{+.052}$ & {\color{black!30} $-0.02_{-.037}^{+.037}$ } & {\color{black!30} $0.06_{-.060}^{+.060}$ } & {\color{black!30} $0.02_{-.062}^{+.062}$ } \\
& \textsc{GPT-3.5-sft-400*} & $1.11_{-.090}^{+.087}$ & $-0.95_{-.075}^{+.075}$ & $-0.30_{-.060}^{+.060}$ & $-0.66_{-.065}^{+.068}$ & $0.76_{-.057}^{+.055}$ & {\color{black!30} $-0.03_{-.042}^{+.045}$ } & $-0.18_{-.052}^{+.052}$ & $-0.23_{-.052}^{+.050}$ \\
& \textsc{GPT-3.5-sft-800*} & $1.02_{-.085}^{+.085}$ & $-0.83_{-.080}^{+.080}$ & $-0.23_{-.067}^{+.065}$ & $-0.55_{-.070}^{+.070}$ & $0.94_{-.055}^{+.055}$ & {\color{black!30} $-0.03_{-.050}^{+.048}$ } & $-0.14_{-.055}^{+.052}$ & $-0.23_{-.062}^{+.060}$ \\
\bottomrule
\end{tabular}
}
    \caption{Mean score differences between original and counterfactual groups, with scores ranging from 1 to 5. Results are shown for both full and stratified subsets (stratified results are marked with *). Subscripts and superscripts indicate confidence intervals, obtained through 10,000 bootstrap iterations. {\color{black!30} Gray values} indicate non-significant differences ($p > 0.01$), while {\color{cdel} coral values} represent significant differences contrary to the expected intervention trend. $(+)$ and $(-)$ denote the expected direction of intervention effect.}
    \label{tab:overall_cfact_results}
    \vspace{-0.3cm}
\end{table*}

\textbf{Firstly}, BERT-like models show higher scoring agreement with human raters than LLMs.
These models can discern complex concepts, i.e., they are highly sensitive to features in the conventions and language complexity categories. After interventions on these concepts, the absolute value of the score difference predicted by BERT-like models often exceeds that of LLMs.
This differs from the phenomenon observed by \citet{singla2023automatic}, in which BERT-based models were found to function like \textit{bag-of-words} models when scoring essays.
However, BERT-like models struggle to distinguish interventions on organization and development. Their predicted score differences are often an order of magnitude smaller than those of LLMs (especially GPT-series models), indicating insensitivity to logical structures and coherence within essays.

\textbf{Secondly}, LLMs respond more comprehensively to our interventions than BERT-like models do, indicating a closer alignment with the criteria specified in scoring rubrics.
As shown in \Cref{tab:model_performance}, LLMs have lower agreement with human ratings when scoring essays in the ZSL setting. However, introducing FSL and SFT considerably improves their performance while maintaining the strength of their rationale alignment, as demonstrated in \Cref{tab:overall_cfact_results}.

\begin{figure}[!h]
    \centering
    \setlength{\abovecaptionskip}{0.cm} 
    \includegraphics[width=\linewidth]{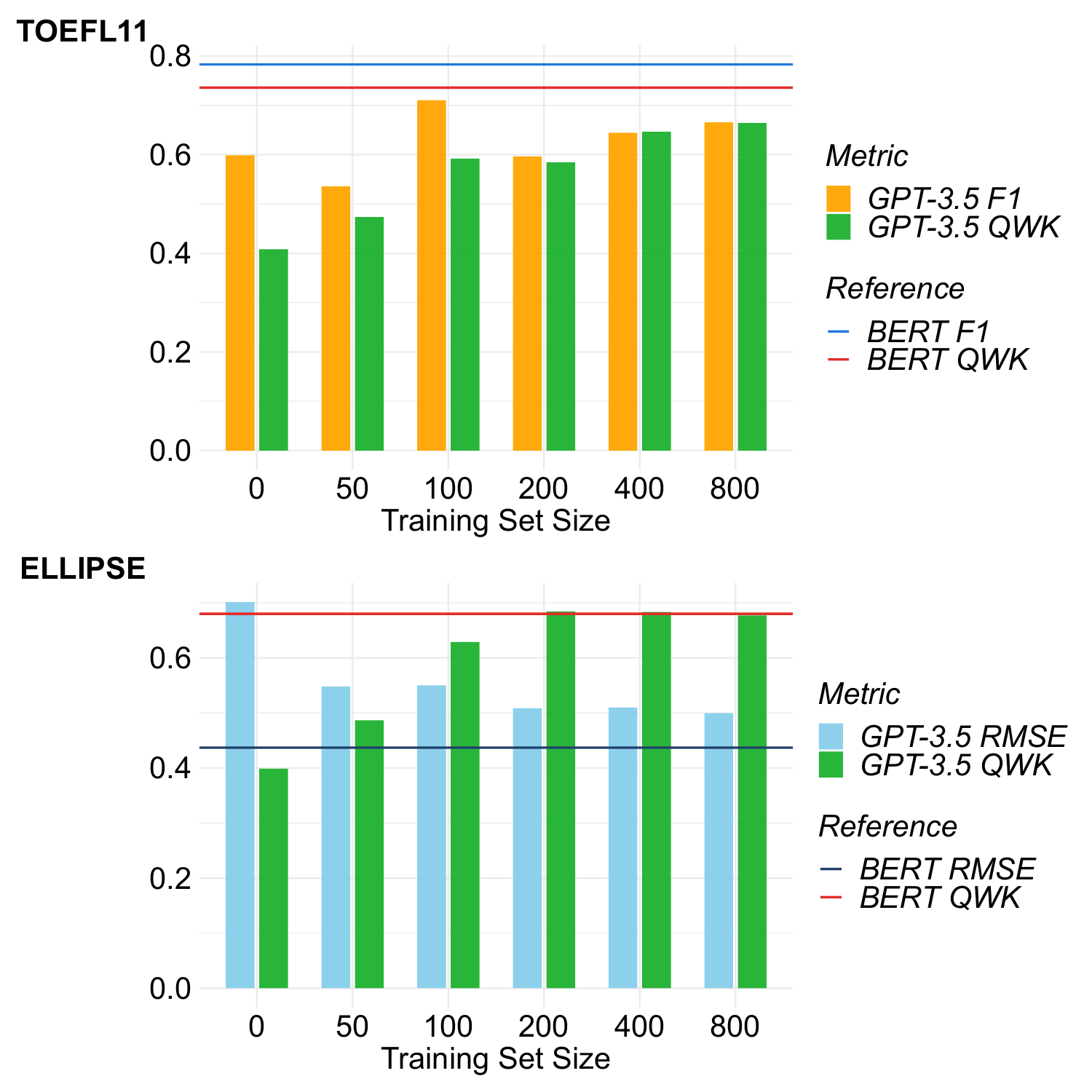}
    \caption{Scoring performance of GPT-3.5 SFT models with varying size of training data. The models' performance improves as the number of training samples increases, reaching comparable or equivalent levels to BERT-like models.}
    \label{fig:ell_ft_train_size}
    \vspace{-0.5cm}
\end{figure}

To better understand the impact of SFT on LLM performance, we fine-tuned GPT-3.5 Turbo with varying data sizes (50, 200, 400, and 800 samples), building on previous results with \textsc{GPT-3.5-sft-100}. 
\Cref{fig:ell_ft_train_size} shows that scoring performance improves as the number of training essays increases. Additionally, with 400 essays for TOEFL11 and 200 for ELLIPSE, the performance nearly stabilizes, reaching levels comparable to BERT, while consistently maintaining rationale alignment capability.

\subsection{Feedback Analysis}
\label{subsec:feedback_provision_exp}

\citet{han_fabricautomatedscoring_2023} and \citet{xiao_automationaugmentationlarge_2024} have proposed that LLMs can provide helpful essay feedback for both writers and evaluators. More importantly, but still largely overlooked, this feedback offers an opportunity to assess the construct validity of models.
Therefore we further investigated the feedback differences provided by LLMs regarding the interventions.
As \Cref{tab:overall_cfact_results} shows that \textsc{gpt-4-fsl} exhibits comprehensive sensitivity in all types of our interventions, while other models typically show inadequate sensitivity in one way or another, we generated feedback using GPT-4 Turbo for further analysis of the model's faithfulness.

As shown in \Cref{fig:feedback_dialogue_example}, after few-shot prompting on scoring task, we continued to ask GPT-4 Turbo to generate feedback based on the scoring rubrics, explaining the scores in terms of aforementioned three concepts.
In this way, we obtained feedback for each essay and its counterfactual counterparts\footnote{We conducted stratified sampling on the ELLIPSE dataset to obtain 200 essay samples and, through two rounds of dialogues, acquired 200 "original-counterfactual" feedback pairs for human evaluation. For the evaluation process, we categorized these pairs based on eight counterfactual interventions and assessed each category of cases accordingly.}. 
Then, three trained annotators were hired to evaluate the feedback differences within each feedback pair, determining whether counterfactual interventions can be detected without accessing essay content.
See the detailed evaluation procedures in \Cref{app:sec:feedback_detail}.

\begin{chat}[!h]
\setlength{\abovecaptionskip}{0.cm} 
\scriptsize \sffamily
\session{\footnotesize Session 1: Essay Scoring}
\user{\underline{\textbf{User:}} Read and evaluate the essay: $\ldots$}
\model{\underline{\textbf{Assistant:}} \{'score': 3.0\}}
\session{\footnotesize Session 2: Providing Feedback}
\user{
\underline{\textbf{User:}} Please provide balanced and constructive feedback on the following aspects of the essay you have just rated (not the example essay):\\
1. Organization:
$\ldots$\\
2. Language Use:
$\ldots$\\
3. Conventions:
$\ldots$\\
Your response should be a structured JSON object with the following keys:\\
\`{}\`{}\`{}
\{\{\\
\quad"organization\_feedback": "",\\
\quad"language\_use\_feedback": "",\\
\quad"conventions\_feedback": ""\\
\}\}
\`{}\`{}\`{}\\
If possible, include direct citations from the essay to substantiate your feedback.
}
\caption{An Example of Feedback Generation}
\label{fig:feedback_dialogue_example}
\vspace{-0.2cm}
\end{chat}

\Cref{tab:feedback_quality_stat} presents the annotator-voted results, demonstrating that a large proportion of counterfactual interventions can be identified simply from the feedback given by the GPT-4 Turbo, especially for complexification, error correction and error introduction except for SVA. On the other hand, simplification and orgnazition interventions are hard to be detected simply from feedback pairs, which is consistent with their relatively smaller absolute effect as shown in \Cref{tab:overall_cfact_results}. One possible reason is that the ELLIPSE essays, written by 8th to 12th grade English learners, tend to be simple in vocabulary and syntax, contain some spelling and SVA errors, and exhibit imperfect logic flow and coherence. Consequently, the model frequently identified SVA issues and offered numerous organizational and developmental suggestions both in feedback of original and counterfactual essays, leading to less distinct differences.

\begin{table}[t!]
    \centering
    \setlength{\abovecaptionskip}{0.1cm}
    \resizebox{\columnwidth}{!}{\begin{tabular}{llc}
\hline
\textbf{Category} & \textbf{Counterfactual Type} & \textbf{Detection Rate\%} \\
\hline
\multirow{4}{*}{Conventions} & Error Correction & 72 \\
                            & Spelling         & 68 \\
                            & SVA              & 48 \\
                            & WOS              & 80 \\
\hline
\multirow{2}{*}{\makecell{Language\\Complexity}}  & Complexification & 100 \\
                                & Simplification   & 32 \\
\hline
\multirow{2}{*}{Organization} & InParaShuffle    & 40 \\
                            & InTextShuffle    & 20 \\
\hline
\end{tabular}
    
}
    \caption{Voting-Based Detection Rates of Original vs. Counterfactual Feedback.}
    \label{tab:feedback_quality_stat}
    \vspace{-0.3cm}
\end{table}

\section{Conclusion}
\label{sec:conclusion}

We generated linguistically-informed counterfactuals with an integrated approach combining LLM and rule-based methods, analyzing their impact on essay scoring results of BERT-like models and LLMs.
Our findings emphasize that a higher scoring agreement with human raters does not necessarily indicate a better alignment with scoring rubrics, suggesting that a more holistic evaluation approach should consider both aspects.
Moreover, our study highlights LLMs' considerable potential in AES domain: while zero-shot prompted LLMs show lower scoring agreement compared to BERT-like models, a major reason for this is that them tend to be conservative or strict when evaluating the essay. FSL and SFT could significantly increase the agreement level with annotated essays serve as anchors to neutralize the conservatism.
In the mean time, LLMs demonstrate comprehensive rationale alignment with scoring rubrics. This ability is stably maintained in ZSL, FSL and SFT settings.
Lastly, LLMs are not only sensitive to counterfactual interventions when scoring but can also reflect a large part of these differences in their feedback, an advantage beyond the reach of traditional AES methods.

This study sheds light on \textcolor{cdel}{\it why} a neural model assigns specific scores to essays. 
It unveils how modifying domain-specific concepts in texts to craft counterfactuals enhances transparency in model decisions—a method applicable across multiple fields. 
With LLMs, counterfactual generation has been greatly empowered, boosting transparency and accountability in machine learning applications.

\newpage
\section{Limitations}
\label{sec:limitations}

In addition to conventions, language complexity, and organization, TOEFL independent writing rubrics also emphasize content-related evaluations—namely, assessing relevance to the prompt and fulfillment of task requirements. These aspects, being beyond mere linguistic concepts, were not included in the current scope of our study. This is because counterfactual interventions require modifying a specific aspect while keeping others constant. This is also because we can adjust linguistic features without affecting content, but altering content inevitably impacts the linguistic aspect. However, we acknowledge that task and topic relevance, as important scoring dimensions, warrant future in-depth exploration.

Our experiment demonstrated that LLMs have significant potential in providing feedback. In this paper, we focus on the feedback differences between original and counterfactual samples. A comprehensive evaluation of the LLM-genearated feedback is a crucial step for future research.

\section*{Acknowledgements}

This research was supported by a grant from the Center for Language Education and Cooperation at the Ministry of Education of China (No. 22YH04ZW), a grant from the National Language Commission of China (No. ZDA145-9), and a grant from the Beijing Federation of Social Science Circles (No. 21DTR037). It is also supported by the Fundamental Research Funds for the Central Universities of China.

\appendix
\clearpage
\centerline{\textbf{\Large Appendix}}
\section{Rubrics}
\label{app:rubric}

To identify the core concepts for intervention, we reviewed five scoring rubrics from IELTS Writing, TOEFL iBT Independent Writing, TOEIC Writing, PTE Academic Writing and the ELLIPSE dataset. We aimed to uncover commonalities across the five rubrics and found that they could be categorized into three dimensions: \textcolor{beautifulblue}{(1)} conventions and accuracy; \textcolor{cdel}{(2)} language complexity; and \textcolor{cadd}{(3)} organization and development.
For clarity, this section will present the descriptors for the highest score in each rubric, with \textbf{color-coded highlights} to indicate the corresponding dimensions.

\subsection{IELTS Writing}

\begin{itemize}
\item Task achievement: fully satisfies all the requirements of the task; clearly presents \textcolor{cadd}{a fully developed response}.

\item Coherence and cohesion: uses \textcolor{cadd}{cohesion} in such a way that it attracts no attention; skillfully \textcolor{cadd}{manages paragraphing}.

\item Lexical resource: uses \textcolor{cdel}{a wide range of vocabulary with very natural and sophisticated control of lexical features}; \textcolor{beautifulblue}{rare minor errors occur only as 'slips'}.

\item Grammatical range and accuracy: uses \textcolor{cdel}{a wide range of structures with full flexibility} and \textcolor{beautifulblue}{accuracy}; \textcolor{beautifulblue}{rare minor errors occur only as 'slips'}.
\end{itemize}

\subsection{TOFEL Independent Writing}
\begin{itemize}
\item Effectively addresses the topic and task.
\item Is \textcolor{cadd}{well organized and well developed}, using clearly appropriate explanations, exemplifications and/or details. 
\item Displays \textcolor{cadd}{unity, progression and coherence}.
\item Displays consistent facility in the use of language, demonstrating \textcolor{cdel}{syntactic variety, appropriate word choice and idiomaticity}, though it may have \textcolor{beautifulblue}{minor lexical or grammatical errors}.
\end{itemize} 

\subsection{TOEIC Writing}
\begin{itemize}  
\item Typically, test takers at level 9 can communicate straightforward information effectively and use reasons, examples, or explanations to support an opinion.
\item When using reasons, examples, or explanations to support an opinion, their writing is \textcolor{cadd}{well-organized and well-developed}.
\item The use of English is natural, with \textcolor{cdel}{a variety of sentence structures and appropriate word choice}, and is \textcolor{beautifulblue}{grammatically accurate}. 
\item When giving straightforward information, asking questions, giving instructions, or making requests, their writing is clear, \textcolor{cadd}{coherent}, and effective.
\end{itemize} 

\subsection{PTE Academic Writing}

\begin{itemize}
\item Content: Adequately deals with the prompt.

\item Form: Length is between 200 and 300 words.

\item Development, Structure \& Coherence: Shows \textcolor{cadd}{good development and logical structure}.

\item Grammar: Shows consistent \textcolor{cdel}{grammatical control of complex language}. \textcolor{beautifulblue}{Errors are rare and difficult to spot}.

\item General Linguistic Range: Exhibits mastery of \textcolor{cdel}{a wide range of language} to formulate thoughts precisely, give emphasis, differentiate and eliminate ambiguity. No sign that the test taker is restricted in what they want to communicate. 

\item Vocabulary: Good command of \textcolor{cdel}{a broad lexical repertoire, idiomatic expressions and colloquialisms}.

\item Spelling: \textcolor{beautifulblue}{Correct spelling}. 
\end{itemize}

\subsection{ELLIPSE Dataset}

\begin{itemize}

\item Overall: Native-like facility in the use of language with \textcolor{cdel}{syntactic variety,
Appropriate word choice and phrases}; \textcolor{cadd}{well-controlled text organization}; \textcolor{beautifulblue}{precise use of grammar and conventions}; \textcolor{beautifulblue}{rare language inaccuracies} that do not impede communication.

\item Cohesion: \textcolor{cadd}{Text organization consistently well controlled using a variety of effective linguistic features such as reference and transitional words and phrases to connect ideas across sentences and paragraphs; appropriate overlap of ideas}.	

\item Syntax: Flexible and effective use of \textcolor{cdel}{a full range of syntactic structures including simple, compound, and complex sentences}; There may be \textcolor{beautifulblue}{rare minor and negligible errors} in sentence formation.	

\item Vocabulary: \textcolor{cdel}{Wide range of vocabulary} flexibly and effectively used to convey precise meanings; skillful use of \textcolor{cdel}{topic-related terms and less common words}; \textcolor{beautifulblue}{rare negligible inaccuracies in word use}.

\item Phraseology: Flexible and effective use of \textcolor{cdel}{a variety of phrases, such as idioms, collocations, and lexical bundles}, to convey precise and subtle meanings; \textcolor{beautifulblue}{rare minor inaccuracies that are negligible}.	  

\item Grammar: Command of grammar and \textcolor{beautifulblue}{usage with few or no errors}.	

\item Conventions: \textcolor{beautifulblue}{Consistent use of appropriate conventions to convey meaning; spelling, capitalization, and punctuation errors nonexistent or negligible}.

\end{itemize}

\section{Detail of Counterfactual Generation}
\label{app:sec:cfact_gen_detail}
In this section, we present the details of our counterfactual generation experiment. This includes examples of rule-based counterfactuals, information on the models used, the prompts provided to the LLMs, and a comparative analysis of various aspects of interest in the counterfactuals generated by GPT-4 Turbo and Llama-3-70b-Instruct.

\subsection{Examples of Rule-based Counterfactuals}
\label{app:subsec:rule_based_cf_examples}

In this study, all the interventions designed to introduce errors into essays and decrease organization are rule-based. In this subsection, we provide counterfactual examples for each of the rule-based interventions. See \Cref{{app:tab:cfact_example_rule}}.

\begin{table*}[ht]
\centering
\setlength{\abovecaptionskip}{0.1cm}
\scriptsize
\resizebox{\linewidth}{!}{
\begin{tabular}{>{\sffamily\raggedright\arraybackslash}m{15cm}}
\toprule
\textbf{Original:}\quad
In the \textcolor{beautifulblue}{up coming} years \textcolor{beautifulblue}{i} \textcolor{beautifulblue}{wanna} save up and work hard to \textcolor{beautifulblue}{get} money, so that \textcolor{beautifulblue}{i} can open my own personal store.\textcolor{beautifulblue}{ .}
\textbackslash n\textbackslash n
\textcolor{beautifulblue}{in} the store \textcolor{beautifulblue}{i} \textcolor{beautifulblue}{going} to \textcolor{beautifulblue}{be} selling high \textcolor{beautifulblue}{branded} shoes\textcolor{beautifulblue}{,}clothes\textcolor{beautifulblue}{,} and \textcolor{beautifulblue}{outer collectives}. \textcolor{beautifulblue}{and} \textcolor{beautifulblue}{Im} going to \textcolor{beautifulblue}{learning} how to invest my money as \textcolor{beautifulblue}{i} sell things before opening up my dream store.
\textbackslash n\textbackslash n
\textcolor{beautifulblue}{i wanna} do this as a job because \textcolor{beautifulblue}{ive awlays had a passhine} for shoes and high \textcolor{beautifulblue}{branded} things. \textcolor{beautifulblue}{and if i} can pull this off \textcolor{beautifulblue}{i think ill be veary succes full} in life \textcolor{beautifulblue}{i will be getting and veary good salary ill be able to pay off anythig and have to money} to support my family.
 \\
\midrule
\textbf{Spelling Error Introduction:}\quad 
In the up coming years i wanna save up and work hard to get money, so that i can open my own personal store.
\textbackslash n\textbackslash n
\textcolor{cdel}{ia} the store \textcolor{cdel}{ye} \textcolor{cdel}{goind} to be \textcolor{cdel}{salling} high branded shoes, \textcolor{cdel}{couthes}, \textcolor{cdel}{anso} outer collectives. and Im going to learning how to invest my money as i sell things before opening up my dream store.
\textbackslash n\textbackslash n
i wanna do \textcolor{cdel}{these} as a job because \textcolor{cdel}{live awlays} had \textcolor{cdel}{e} passhine \textcolor{cdel}{foi} \textcolor{cdel}{shooes} and \textcolor{cdel}{highest} branded things. and if i can pull this off i think ill be veary succes full in life i will be getting and veary good salary ill be able to pay off anythig and have to money to support my family.
 \\
\midrule
\textbf{Subject-verb Agreement Error Introduction:}\quad
In the up coming years i wanna save up and work hard to get money, so that i can open my own personal store .
\textbackslash n\textbackslash n
in the store i going to be selling high branded shoes,clothes, and outer collectives. and Im going to learning how to invest my money as i sell things before opening up my dream store.
\textbackslash n\textbackslash n
i wanna \textcolor{cdel}{does} this as a job because ives awlays had a passhine for shoes and high branded things. and if i can pull this off i thinks ill is veary succes full in life i will be getting and veary good salary ill be able to pay off anythig and have to money to support my family.
 \\
\midrule
\textbf{Word Order Swapping:}\quad 
In the up coming years i wanna save up and work hard to get money, so that i can open my own personal store .
\textbackslash n\textbackslash n
\textcolor{cdel}{the in store i be going to selling branded high}, shoes clothes, and collectives outer. and Im going to learning how to invest my money as i sell things before opening up my dream store.
\textbackslash n\textbackslash n
i wanna do this as a job because ive awlays had a passhine for shoes and high branded things. and if i can pull off this i ill think be veary succes full in \textcolor{cdel}{i life be will getting and veary salary good} ill be able to pay off anythig and to have money to support my family.
 \\
\midrule
\textbf{Intra-paragraph Shuffling:}\quad 
In the up coming years i wanna save up and work hard to get money, so that i can open my own personal store .
\textbackslash n\textbackslash n
\textcolor{cadd}{and Im going to learning how to invest my money as i sell things before opening up my dream store. in the store i going to be selling high branded shoes,clothes, and outer collectives.}
\textbackslash n\textbackslash n
and if i can pull this off i think ill be veary succes full in life i will be getting and veary good salary ill be able to pay off anythig and have to money to support my family. i wanna do this as a job because ive awlays had a passhine for shoes and high branded things.
 \\
\midrule
\textbf{Inter-text Shuffling:}\quad 
In the up coming years i wanna save up and work hard to get money, so that i can open my own personal store .
\textbackslash n\textbackslash n
\textcolor{cadd}{i wanna do this as a job because ive awlays had a passhine for shoes and high branded things.and Im going to learning how to invest my money as i sell things before opening up my dream store.
\textbackslash n\textbackslash n
in the store i going to be selling high branded shoes,clothes, and outer collectives.}and if i can pull this off i think ill be veary succes full in life i will be getting and veary good salary ill be able to pay off anythig and have to money to support my family.
 \\
\bottomrule
\end{tabular}
}
\caption{Example of the medium-level TOEFL11 essay used in \Cref{tab:cfact_example} and its rule-based counterfactual counterparts.}
\label{app:tab:cfact_example_rule}
\vspace{-0.5cm}
\end{table*}

\subsection{Prompts for Counterfactual Generation}
\label{app:subsec:cfact_gen_prompt}

The counterfactual samples of text correction, complexification and simplification are generated by the \texttt{gpt-4-1106-preview} model.
When calling OpenAI's APIs, we turn on \texttt{JSON} mode to get easier parsing results. For reproducibility, we set the \texttt{temperature} parameter to 0 and the \texttt{seed} to 42.

\subsubsection{Prompt for Error Correction}

\begin{tcolorbox}[breakable,title=\textbf{System:} You are an experienced writing tutor.]
\underline{\textbf{User:}} Please fix the spelling, punctuation and grammatical errors in the given essay. Ensure the main idea, the words used, the sentence structure, and the length of the text remain consistent with the original text.

\vspace{0.5cm}

Input Essay:\\
"\{\}"

\vspace{0.5cm}

Please return the output essay in JSON format as below:

\`{}\`{}\`{}\\
\texttt{\{"output\_essay": "..."\}}\\
\`{}\`{}\`{}

Output:
\end{tcolorbox}

\subsubsection{Prompt for Complexification}

\begin{tcolorbox}[breakable,title=\textbf{System:} You are an experienced writing tutor.]
\underline{\textbf{User:}} Modify the provided essay to enhance its lexical sophistication and syntactic variety following the instructions below:  

1. Expand lexical range: Vary word choice and replace common words with advanced vocabulary when suitable without compromising clarity or meaning. Avoid repeating the same words and capture subtle differences in meaning.  

2. Increase syntactic complexity: Incorporate a wider range of sentence structures including compound-complex sentences, varied clause types, subordination and coordination. Use advanced constructions such as non-finite clauses, adverbials, conditionals, inversion and passives where appropriate.

3. Maintain meaning, length and clarity: The revised text should retain the original ideas and conform to the initial length while remaining clear and understandable.

\vspace{0.5cm}

Input Essay:\\
"\{\}"

\vspace{0.5cm}

Please return the output essay in JSON format as below: 

\`{}\`{}\`{}\\
\texttt{\{"output\_essay": "..."\}}\\
\`{}\`{}\`{}

Output:
\end{tcolorbox}

\subsubsection{Prompt for Simplification}

\begin{tcolorbox}[breakable,title=\textbf{System:} You are an experienced writing tutor.]
\underline{\textbf{User:}} Modify the provided essay to simplify its vocabulary and sentence structure following the instructions below:

1. Simplify vocabulary: Replace advanced words with common everyday equivalents for clear understanding. Limit synonyms to favor those most commonly used.

2. Simplify sentence structure: Break down complex sentences and avoid clauses, conjunctions, and nesting where possible. Favor short, simple subject-verb-object sentences.

3. Maintain meaning, length and clarity: The revised text should retain the original ideas and conform to the initial length while remaining clear and understandable.

\vspace{0.5cm}

Input Essay:\\
"\{\}"

\vspace{0.5cm}

Please return the output essay in JSON format as below:\\
\`{}\`{}\`{}\\
\texttt{\{"output\_essay": "..."\}}\\
\`{}\`{}\`{}

Output:
\end{tcolorbox}

\subsection{Comparative Performance of GPT-4 Turbo and Llama-3-70b-Instruct in Counterfactual Generation}
\label{app:subsec:compare_gpt4_llama_cf_gen}

\Cref{app:tab:cfact_ling_gpt4_llama} shows the effect size of three types of interventions performed by both GPT-4 Turbo and Llama-3-70b-Instruct on seven linguistic metrics across two datasets.
It can be seen that the impact of the two models on the original essay, across various language metrics of interest during counterfactual interventions, aligns with expectations, albeit with slight variations in degree. In terms of error correction, GPT-4 significantly reduces error density. Meanwhile, for complexification and simplification, GPT-4 intervenes more comprehensively in vocabulary and syntax, with generally smaller changes in length.

\Cref{tab:cos_sim_gpt4_cf} presents the embedding similarities bewteen counterfactuals and original essays given by both LLMs.
Although Llama-3-70b-Instruct retains a higher degree of the original text's meaning than GPT-4 Turbo in most cases, it shows a significant drop when simplifying the ELLIPSE essay, indicating its potential lack of stability.

\begin{table*}[h!]
    \centering
    \resizebox{\linewidth}{!}{\begin{tabular}[t]{cccccccc}
\toprule
\multirow{2}{*}{Metrics} & \multirow{2}{*}{Model} & \multicolumn{3}{c}{\textbf{TOEFL11}} & \multicolumn{3}{c}{\textbf{ELLIPSE}} \\
\cmidrule(r){3-5} \cmidrule(r){6-8}
& & \textbf{Error Correction} & \textbf{Complexification} & \textbf{Simplification} & 
\textbf{Error Correction} & \textbf{Complexification} & \textbf{Simplification}\\
\midrule
\multirow{2}{*}{\tt \tt WordNum} & \textsc{Llama-3-70b-It}
 & -0.045  & -0.170  & -1.325  & 0.065  & -0.332  & -0.981 \\
 & \textsc{gpt-4 Turbo}
 & -0.098  & 0.078  & -1.060  & -0.027  & -0.103  & -0.714 \\
\midrule
\multirow{2}{*}{\tt SentNum} & \textsc{Llama-3-70b-It}
 & 0.037  & -0.508  & -0.037  & 0.215  & -0.406  & -0.074 \\
 & \textsc{gpt-4 Turbo}
 & 0.047  & -0.323  & 0.454  & 0.280  & -0.264  & 0.367 \\
\midrule
\multirow{2}{*}{\tt MLS} & \textsc{Llama-3-70b-It}
 & -0.176  & 0.385  & -1.473  & -0.354  & 0.156  & -1.816 \\
 & \textsc{gpt-4 Turbo}
 & -0.245  & 0.449  & -1.714  & -0.481  & 0.423  & -2.237 \\
\midrule
\multirow{2}{*}{\tt ADDT} & \textsc{Llama-3-70b-It}
 & -0.030  & 0.734  & -1.359  & -0.353  & 0.817  & -1.628 \\
 & \textsc{gpt-4 Turbo}
 & -0.066  & 0.982  & -1.535  & -0.481  & 1.220  & -1.875 \\
\midrule
\multirow{2}{*}{\tt LemmaTTR} & \textsc{Llama-3-70b-It}
 & -0.074  & 2.130  & -0.467  & 0.020  & 2.647  & -0.009 \\
 & \textsc{gpt-4 Turbo}
 & 0.094  & 2.985  & -0.611  & 0.429  & 3.323  & -0.128 \\
\midrule
\multirow{2}{*}{\tt LexSoph} & \textsc{Llama-3-70b-It}
 & -1.538  & 3.596  & -0.186  & -0.799  & 3.710  & 0.301 \\
 & \textsc{gpt-4 Turbo}
 & -1.514  & 5.277  & -0.909  & -0.711  & 5.063  & -0.291 \\
\midrule
\multirow{2}{*}{\tt ErrorDensity} & \textsc{Llama-3-70b-It}
 & -5.015  & -0.616  & -0.535  & -1.887  & -0.628  & -0.331 \\
 & \textsc{gpt-4 Turbo}
 & -5.122  & -0.407  & -0.219  & -1.869  & -0.412  & -0.123 \\
\bottomrule
\end{tabular}
}
    \caption{Cohen's $\mathcal{D}$ for seven linguistic metrics on three interventions of GPT-4 Turbo and Llama-3-70b-Instruct.}
    \label{app:tab:cfact_ling_gpt4_llama}
\end{table*}

\begin{table}[h!]
    \centering
    \resizebox{0.98\columnwidth}{!}{\begin{tabular}[t]{lccc}
\toprule
Model & Intervention & TOEFL11 & ELLIPSE \\
\midrule
\multirow{3}{*}{\textsc{gpt-4 turbo}} &
Error Correction & 0.935 & 0.942 \\
& Complexification & 0.760 & 0.749 \\
& Simplification & 0.816 & 0.849 \\
\midrule
\multirow{3}{*}{\textsc{Llama-3-70b-it}} &
Error Correction & 0.944 & 0.957 \\
& Complexification & 0.817 & 0.813 \\
& Simplification & 0.853 & \textcolor{cdel}{0.610} \\
\bottomrule
\end{tabular}
}
    \caption{Mean cosine similarity between original and counterfactual essays for GPT-4 and Llama-3-70b-instruct given by \texttt{text-embedding-3-large}.}
    \label{app:tab:cos_sim_cf_gpt4_llama3_70b}
\end{table}

\section{The Implementation of AES methods}
\label{app:sec:scoring_implement_detail}

\subsection{Fine-tuning BERT-like Models}

We fine-tuned three commonly used pre-trained transformer-based encoder models, specifically \texttt{bert-base-uncased}, \texttt{roberta-base}, and \texttt{deberta-v3-base}.

\subsubsection{Basic Settings}

As the essays in the TOEFL11 dataset are categorized into low, medium, and high categories, we developed a three-class classifier using the cross-entropy loss.
We use the \texttt{AutoModelForSequenceClassification} class from Hugging Face \texttt{transformer}, setting \texttt{num\_labels=3} to load the pre-training checkpoints. 
For the ELLIPSE dataset, with scores ranging from 1.0 to 5.0, we model it as a regression problem by setting \texttt{num\_labels=1} and using the mean squared error (MSE) loss function.

\subsubsection{Hyperparameters}

In our model fine-tuning process, we experimented with four distinct learning rates: 1e-5, 2e-5, 3e-5, and 5e-5, using Hugging Face's \texttt{Trainer}. We identify the best learning rate that led to the lowest loss on the validation set (results see \Cref{app:tab:lr_tuning}). 
We used a linear learning rate scheduler that includes a 50-step warm-up phase, where the learning rate initially increases from a lower value to a specified maximum (chosen from the four rates: 1e-5, 2e-5, 3e-5, and 5e-5) and then decreases linearly. This method ensures gradual adaptation of the model's weights, with the peak learning rates being reached at the end of the warm-up.

For other parameters, we used a seed of 42 and a batch size of 16 for both training and evaluation. We aimed for a maximum of 10 epochs, with the actual duration potentially reduced by early stopping, triggered if loss value fails to improve after 5 checks. The approach included a weight decay of 0.01 for overfitting prevention and \texttt{FP16} for efficient training. Input lengths were adjusted to 512 tokens through padding and truncation to ensure uniformity across all samples.

\begin{table}[t]
\centering
    \resizebox{\linewidth}{!}{\begin{tabular}{ccccc}
\hline
\textbf{Dataset} & \textbf{Model} & \textbf{Learning Rate} & \textbf{EarlyStop@Step} & \textbf{Validation Loss} $\downarrow$ \\
\hline
\multirow{12}{*}{TOEFL11} & \multirow{4}{*}{\textsc{BERT}} & \textbf{1e-5} & \textbf{450} & \textbf{.443} \\
&                           & 2e-5        & 550 &.453 \\
&                           & 3e-5        & 350 & .462 \\
&                           & 5e-5        & 150 & .482 \\
\cmidrule(r){2-5}
& \multirow{4}{*}{\textsc{RoBERTa}}  & \textbf{1e-5} & \textbf{450} & \textbf{.403} \\
&                           & 2e-5         & 450 & .424 \\
&                           & 3e-5         & 400 & .442 \\
&                           & 5e-5         & 500 & .467 \\
\cmidrule(r){2-5}
& \multirow{4}{*}{\textsc{DeBERTa}}  & \textbf{1e-5} & \textbf{500} & \textbf{.398} \\
&                           & 2e-5         & 400 & .400 \\
&                           & 3e-5              & 250 & .416 \\
&                           & 5e-5              & 250 & .427 \\
\hline
\multirow{12}{*}{ELLIPSE} & \multirow{4}{*}{\textsc{BERT}} & 1e-5 & 500 & .173 \\
&                           & \textbf{2e-5}         & \textbf{200} & \textbf{.172} \\
&                           & 3e-5              & 300 & .179 \\
&                           & 5e-5              & 150 & .185 \\
\cmidrule(r){2-5}
& \multirow{4}{*}{\textsc{RoBERTa}} & 1e-5 & 250 & .196 \\
&                           & 2e-5         & 100 & .199 \\
&                           & \textbf{3e-5}              & \textbf{500} & \textbf{.171} \\
&                           & 5e-5              & 300 & .176 \\
\cmidrule(r){2-5}
& \multirow{4}{*}{\textsc{DeBERTa}}  & \textbf{1e-5} & \textbf{200} & \textbf{.157} \\
&                           & 2e-5         & 150 & .167 \\
&                           & 3e-5              & 200 & .160 \\
&                           & 5e-5              & 150 & .181 \\
\hline
\end{tabular}
    
}
    \caption{Performance of the three models on the validation set after fine-tuning using different learning rates on both TOEFL11 and ELLIPSE datasets. Learning rates for achieving minimum loss in each model for both datasets are \textbf{bolded}.}
    \label{app:tab:lr_tuning}
    \vspace{-0.4cm}
\end{table}

\subsection{Prompting LLMs to Score Essays}

As introduced in \Cref{sec:experiments}, we also used LLMs for essay scoring, including \texttt{gpt-3.5-turbo-1106} and \texttt{gpt-4-1106-preview} based on OpenAI's API. We turned on \texttt{JSON} mode to get easier parsing results, and set the \texttt{temperature} parameter to 0 and the \texttt{seed} parameter to 42 for reproducibility.

\subsubsection{Prompts for Scoring TOEFL11 Essays with Zero-shot Learning}

Below is the scoring template for TOEFL11 essays. In the zero-shot setting, we provide the LLMs with the essay prompt, the essay itself, and the scoring rubrics. Notably, while the TOEFL11 dataset only provides low, medium, and high score levels for the essays without specific scores, the TOEFL rating rubric is actually based on a 1 to 5 scale. Consequently, even in zero-shot scenarios without examples or training data, we can still prompt LLMs to assess and score TOEFL11 essays.

\begin{tcolorbox}[breakable,title=\textbf{System:} You are a TOEFL rater specializing in the evaluation of the Independent Writing section.]
\underline{\textbf{User:}} Read and evaluate the essay written in response to the prompt: "\{\}"

\vspace{0.5cm}

Essay:
"\{\}"

\vspace{0.5cm}

Please assign it a score from 1 to 5 (in increments of 0.5 points) based on rubric below:

"\{TOEFL11\_RUBRICS\}"

\vspace{0.5cm}

Your response should be a JSON object containing only one key: 'score', which should be a numeric value representing the score you gave.
\end{tcolorbox}

\begin{tcolorbox}[breakable,colback=red!5!white,colframe=red!75!black,title=\textbf{TOEFL11 Rubrics}] 
- A 5-point essay effectively addresses all aspects of the topic and task. Well organized and developed with clearly appropriate explanations and details. Displays strong unity, progression and coherence. Shows consistent language facility with syntactic variety, appropriate word choice and idiomaticity. May have minor lexical or grammatical errors.

- A 4-point essay addresses the topic and task well, though some points may not be fully elaborated. Generally well organized and developed with appropriate and sufficient explanations, exemplifications and details. Displays unity, progression and coherence, though may contain occasional redundancy, digression or unclear connections. Demonstrates syntactic variety and vocabulary range. May have occasional minor errors that do not interfere with meaning.

- A 3-point essay addresses the topic and task with somewhat developed explanations, exemplifications and details. Displays unity, progression and coherence, though connection of ideas may be occasionally obscured. May demonstrate inconsistent language facility resulting in lack of clarity and obscured meaning. May display accurate but limited structures and vocabulary.

- A 2-point essay shows limited development in response to the topic and task. Inadequate organization or connection of ideas. Insufficient or inappropriate exemplifications, explanations or details to support generalizations. Noticeable inappropriate word choices or word forms. An accumulation of errors in sentence structure and/or usage.

- A 1-point essay is seriously flawed due to disorganization, underdevelopment, little or no supporting detail, and unresponsiveness to the task. Contains serious and frequent errors in sentence structure or usage.
\end{tcolorbox}

\subsubsection{Prompts for Scoring ELLIPSE Essays with Zero-shot Learning}

Below is the scoring template for ELLIPSE essays. 
Since the ELLIPSE's rubrics do not require adherence to a specific prompt or fulfillment of task requirements. We only provide the LLMs with the essay to be rated and the scoring rubrics. 

\begin{tcolorbox}[breakable,title=\textbf{System:} You are an essay rater specializing in the evaluation of essays written by students from 8th to 12th grade who are learning English as a second language.]
\underline{\textbf{User:}} Read and evaluate the essay:
"\{\}"

\vspace{0.5cm}

Assign it a score from 1 to 5, in increments of 0.5, based on this rubric:

"\{ELLIPSE\_RUBRICS\}"

\vspace{0.5cm}

Your response should be a JSON object containing only one key: 'score', which should be a numeric value representing the score you gave.
\end{tcolorbox}

\begin{tcolorbox}[breakable,colback=red!5!white,colframe=red!75!black,title=\textbf{ELLIPSE Rubrics}] 
- A 5-point essay demonstrates native-like facility in the use of language with syntactic variety, appropriate word choice and phrases; well-controlled text organization; precise use of grammar and conventions; rare language inaccuracies that do not impede communication.

- A 4-point essay demonstrates facility in the use of language with syntactic variety and range of words and phrases; controlled organization; accuracy in grammar and conventions; occasional language inaccuracies that rarely impede communication.

- A 3-point essay demonstrates facility limited to the use of common structures and generic vocabulary; organization generally controlled although connection sometimes absent or unsuccessful; errors in grammar and syntax and usage. Communication is impeded by language inaccuracies in some cases.

- A 2-point essay demonstrates inconsistent facility in sentence formation, word choice, and mechanics; organization partially developed but may be missing or unsuccessful. Communication impeded in many instances by language inaccuracies.

- A 1-point essay demonstrates a limited range of familiar words or phrases loosely strung together; frequent errors in grammar (including syntax) and usage. Communication impeded in most cases by language inaccuracies.
\end{tcolorbox}

\subsubsection{Prompts for Scoring TOEFL11 Essays with Few-shot Learning}

For few-shot learning on TOEFL11 dataset, we gave three examples from the low, medium and high categories, and asked the models to return the score level as well. See the prompt below.

\begin{tcolorbox}[breakable,title=\textbf{System:} You are a TOEFL rater specializing in the evaluation of the Independent Writing section.]
\underline{\textbf{User:}} Read and evaluate the essay written in response to the prompt: "\{\}" 

\dashedline

Example essay 1 of score level "High":\\
"\{{\scriptsize A\_REPRESENTATIVE\_HIGH\_SCORE\_ESSAY}\}"

\dashedline

Example Essay 2 of score level "Medium":\\
"\{{\scriptsize A\_REPRESENTATIVE\_MEDIUM\_SCORE\_ESSAY}\}"

\dashedline

Example Essay 3 of score level "Low":\\
"\{{\scriptsize A\_REPRESENTATIVE\_LOW\_SCORE\_ESSAY}\}"

\dashedline

Essay to score:\\
"\{\}"

\vspace{0.5cm}

Please note:\\
- Low corresponds to scores of 1.0 - 2.0\\
- Medium corresponds to scores of 2.5 - 3.5\\  
- High corresponds to scores of 4.0 - 5.0\\

\vspace{0.5cm}

Assign the essay a score level of Low, Medium or High based on the criteria in the rubric below:

"\{TOEFL11\_RUBRICS\}"

\vspace{0.5cm}

Your response should be a JSON object with the key "\texttt{score\_level}" set to either "\texttt{Low}", "\texttt{Medium}", or "\texttt{High}" representing the level you determined for this essay.
\end{tcolorbox}

\subsubsection{Prompts for Scoring ELLIPSE Essays with Few-shot Learning}

To align with the process of rating TOEFL11 essays, we also provide three example essays from the ELLIPSE dataset, representing low, medium, and high score levels. However, we give the specific scores of these examples and require the model to return numerical scores as well. Refer to the following prompt. For information on how to select samples, see the next section.

\begin{tcolorbox}[breakable,title=\textbf{System:} You are an essay rater specializing in the evaluation of essays written by students from 8th to 12th grade who are learning English as a second language.]
\underline{\textbf{User:}} Read and evaluate the essay: 

\dashedline

Example essay 1 of score "4.0":\\
"\{{\scriptsize A\_REPRESENTATIVE\_HIGH\_SCORE\_ESSAY}\}"

\dashedline

Example Essay 2 of score "3.0":\\
"\{{\scriptsize A\_REPRESENTATIVE\_MEDIUM\_SCORE\_ESSAY}\}"

\dashedline

Example Essay 3 of score "2.0":\\
"\{{\scriptsize A\_REPRESENTATIVE\_LOW\_SCORE\_ESSAY}\}"

\dashedline

Essay to score:\\
"\{\}"

\vspace{0.5cm}

Assign it a score from 1 to 5, in increments of 0.5, based on this rubric:

"\{ELLIPSE\_RUBRICS\}"

\vspace{0.5cm}

Your response should be a JSON object containing only one key: 'score', which should be a numeric value representing the score you gave.
\end{tcolorbox}

\subsubsection{Few-shot Example Selection}

We use a linguistic-based approach to select the representative examples for few-shot learning by following the steps:
\begin{enumerate}
    \item \textbf{Calculate Metrics:} Calculate and normalize the seven linguistic metrics mentioned in \Cref{subsec:validity_cf_gen} for training sets of both TOEFL11 and ELLIPSE datasets.
    \item \textbf{Process Data:} Apply Principal Component Analysis (PCA) to identify the top five components that explain 95\% of the variance, representing essential linguistic metrics.
    \item \textbf{Represent Samples:} Utilize these principal components to represent the linguistic metrics of all training samples.
    \item \textbf{Determine Medoids:} Categorize samples into proficiency levels (low, medium, high) and find the medoid of each group using Euclidean distance.
\end{enumerate}

Note that a medoid is an object within a dataset that minimally differs from all other objects in the dataset, according to a given distance metric. It is similar to the concept of a centroid, but while a centroid may not be an actual data point, a medoid is always a member of the dataset.

\subsection{Fine-tuning GPT-3.5 Turbo}
\label{subsec:ft-gpt-35}

We fine-tuned GPT-3.5 Turbo model using the OpenAI API\footnote{\url{https://platform.openai.com/docs/guides/fine-tuning}} with the following hyperparameters: 3 epochs, a batch size of 1, and a learning rate multiplier of 2. These are the default settings provided by OpenAI, as the size and weight of GPT-3.5 Turbo model are not accessible, a systematic parameter search would be very costly and even impossible.

\section{Details for Feedback Generation and Evaluation}
\label{app:sec:feedback_detail}

\subsection{Feedback Generation}

Given the stable performance of few-shot GPT-4 in handling a variety of counterfactual interventions, we conducted the manual evaluations on this model.
As shown in \Cref{fig:feedback_dialogue_example}, we prompted the few-shot GPT-4 to provide writing feedback to the essay it just scored. The experiments were conducted on a stratified subset of ELLIPSE. For 200 samples in the subset, we requested GPT-4 to provide feedback respectively on each of the original samples and their specific counterfactual counterparts.

\begin{tcolorbox}[breakable,colback=teal!5!white,colframe=teal!75!black,title=\textbf{Full Prompt Instructing GPT-4 to Provide Feedback}]
\underline{\textbf{User:}} Please provide balanced and constructive feedback on the following aspects of the essay you have just rated (not the example essay):

1. Organization: \\
- Evaluate how effectively ideas are communicated and organized. Identify any issues with the logical flow, transitions between ideas, and clarity in conveying concepts. Comment on the introduction's setup, idea development throughout the body, and the conclusiveness of the ending.

2. Language Use: \\
   - Morphology: Identify errors in word formation and structure, focusing on verb tenses, irregular verbs, plurals, possessives, affixes, agreement, and gerund/participle usage.\\
   - Syntax: Comment on the arrangement of words and phrases to create well-formed sentences, coherence in sentence construction, and the complexity and variety of sentence types.\\
   - Vocabulary: Assess the appropriateness of word choice, the diversity and sophistication of vocabulary employed, and note any imprecise use of words where more accurate or specific terms could be used.

3. Conventions: \\
   - Highlight any errors in spelling, capitalization, and punctuation.

Your response should be a structured JSON object with the following keys:\\
\`{}\`{}\`{}json\\
\{\{\\
\quad"\texttt{organization\_feedback}": "",\\
\quad"\texttt{language\_use\_feedback}": "",\\
\quad"\texttt{conventions\_feedback}": ""\\
\}\}\\
\`{}\`{}\`{}\\
If possible, include direct citations from the essay to substantiate your feedback.
\end{tcolorbox}

\subsection{Feedback Evaluation}

After collecting 200 "original-counterfactual" feedback pairs, we divided them into 8 equal portions, with each portion corresponding to one counterfactual intervention for manual evaluation. 
Three graduate students majoring in Linguistics were tasked with identifying feedback differences within each pair by deciding whether one of the feedback better aligned with specific features. For instance, for interventions introducing \textit{spelling errors}, we asked, 
\begin{quote}
    \textit{Which feedback (~\RomanNumeralCaps{1} or \RomanNumeralCaps{2}~) more clearly or frequently points out spelling errors?}
\end{quote}

Annotators could choose from four choices:

\begin{quote}\it
    \begin{enumerate}[label=(\alph*)]
    \item Feedback~\RomanNumeralCaps{1};
    \item Feedback~\RomanNumeralCaps{2};
    \item Both are similar;
    \item Uncertain.
    \end{enumerate}
\end{quote}

Correct identification of the counterfactual feedback was recorded as "correct", incorrect identification was recorded as "incorrect", and choosing options (c) or (d) was recorded as "Indeterminate".

Three graduate students majoring in Linguistics were tasked with identifying the feedback differences within each pair. The final classification for each pair was determined by a majority vote among the annotators. In cases where no majority was reached due to each annotator choosing a different outcome, the pair was labeled as "indeterminate".
Results are in \Cref{app:tab:feedback_eval_vote}. \Cref{app:tab:feedback_example_spelling} gives an example of a feedback pair where the counterfactual feedback corresponds to a sample obtained by introducing spelling errors to the original sample.

\begin{table*}[!t]\small
\centering
    \begin{tabular}{llccc}
\hline
\textbf{Category} & \textbf{Counterfactual Type} & \textbf{Correct\%} & \textbf{Incorrect\%} & \textbf{Indeterminate\%} \\
\hline
\multirow{4}{*}{Conventions \& Accuracy} & Error Correction & 72 & 4 & 24 \\
                            & Spelling         & 68  & 0 & 32\\
                            & SVA              & 48 & 0 & 52\\
                            & WOS              & 80 & 0 & 20 \\
\hline
\multirow{2}{*}{Language Complexity}  & Complexification & 100 & 0 & 0 \\
                                & Simplification   & 32 & 0 & 68 \\
\hline
\multirow{2}{*}{\makecell{Organization \& Development}} & InParaShuffle    & 40 & 4 & 56\\
                            & InTextShuffle    & 20 & 8 & 72 \\
\hline
\end{tabular}

    \caption{Feedback Type Voting Results by Annotators}
    \label{app:tab:feedback_eval_vote}
    \vspace{-0.5cm}
\end{table*}

\begin{table*}[ht!]
\small
\centering
\begin{tabularx}{\textwidth}{@{}X|X@{}}
\toprule
\multicolumn{1}{c|}{Feedback for the Original}                                                                                                                                                                                                                                                                                                                                                                                                                                                    & \multicolumn{1}{c}{Feedback for the Counterfactual}                                                                                                                                                                                                                                                                                                                                                                                                                                                                                                                                                        \\ \midrule
**organization\_feedback**\\
The essay begins with a clear topic sentence that introduces the importance of parks for improving mood, but the development of ideas throughout the body is somewhat disjointed. The writer provides personal anecdotes and general observations to support the argument, but transitions between ideas are abrupt, and there is a lack of clear logical flow. The conclusion does not effectively summarize the main points or provide a strong final thought on the topic. The essay would benefit from a more structured approach, with each paragraph focusing on a single aspect of the importance of parks and using transitions to guide the reader through the argument.

\vspace{0.5cm}

**language\_use\_feedback**

Morphology: There are several errors in verb tense consistency ('i use to just go outside' should be 'I used to just go outside') and subject-verb agreement ('kids finds parks activity enjoyable' should be 'kids find park activities enjoyable'). Syntax: Sentence structures are often simplistic and repetitive, lacking variety and complexity. There are also instances of awkward phrasing ('my mom always straggle with tell my brothers to go outside' should be 'my mom always struggles to tell my brothers to go outside'). Vocabulary: The vocabulary is generally appropriate but lacks diversity and sophistication. There are instances of imprecise language ('be open minded' could be 'become open-minded', 'enjoy life' could be 'enjoy the outdoors').

\vspace{0.5cm}

**conventions\_feedback**

\textcolor{cdel}{The essay contains several spelling errors ('straggle' should be 'struggle', 'rob' should be 'rope')}, incorrect capitalization ('iPod', 'iPhone', 'I', and the first word of sentences should be capitalized), and punctuation mistakes (missing commas in compound sentences, incorrect spacing around punctuation marks). Attention to these details would improve the readability and professionalism of the essay.

 & **organization\_feedback**
 
 The essay's organization needs improvement. Ideas about the importance of parks and their impact on mood and childhood experiences are presented, but the logical flow is disjointed. Transitions between ideas are abrupt or missing, making it difficult to follow the argument. The introduction lacks a clear thesis statement, and the body paragraphs do not consistently develop the initial idea. The conclusion is not clearly defined, and the essay ends abruptly without summarizing the main points or providing a final thought on the importance of parks for children.

 \vspace{0.5cm}
 
 **language\_use\_feedback**
 
 Morphology: There are several errors in verb tenses and subject-verb agreement (e.g., 'my mm takes him ty the park and ha play' should be 'my mom takes him to the park and he plays'). Syntax: Sentence structures are often incomplete or incoherent, which disrupts the flow of information (e.g., 'Know i barely even go outside i go ti school and then i come back is just' is difficult to understand). Vocabulary: Word choice is sometimes inaccurate or too simplistic, and there is a lack of diversity in vocabulary. More precise language could enhance the clarity of the essay (e.g., 'bouring' should be 'boring', 'actvity enjoyble' should be 'activities enjoyable').

 \vspace{0.5cm}
 
 **conventions\_feedback**
 
 \textcolor{cdel}{The essay contains numerous spelling errors (e.g., 'iMy', 'btother', 'ty', 'ha play', 'lf', 'alls', 'jn', 'tie', 'suing', 'actvity', 'mint', 'whi offen', 'leke', 'ext', 'litter', 'bi on m'i phone', 'utside', 'tou', 'weill', 'offen mindend', 'thingh', 'dayl', 'field', 'wak ta giong tte park', 'nown', 'mer', 'estay', 'gou').} Capitalization is inconsistent, and punctuation is often missing or used incorrectly, making it challenging to understand the intended meaning of sentences.
\\ \bottomrule
\end{tabularx}

\caption{Example feedback pair of original and counterfactual obtained by introducing spelling errors.}
\label{app:tab:feedback_example_spelling}
\vspace{-0.5cm}
\end{table*}

\subsection{Ethical Considerations}
The three annotators involved in this project were graduate students in linguistics. Prior to assigning them the annotation task, we provided a comprehensive introduction to the content, purpose, and significance of the project. Each annotator was responsible for reviewing 200 feedback pairs and received compensation of \$0.42 per annotated pair.

\end{document}